\documentclass{article}

\usepackage{microtype}
\usepackage{graphicx}
\usepackage{subfigure}
\usepackage{booktabs} 

\usepackage{listings}
\usepackage{color}
\usepackage{amsmath,amsfonts,bm}

\newcommand{\ie}{i.e.,~}

\newcommand{\diag}[1]{\textbf{diag}\left(#1\right)}

\DeclareMathAlphabet{\mathsfit}{\encodingdefault}{\sfdefault}{m}{sl}
\SetMathAlphabet{\mathsfit}{bold}{\encodingdefault}{\sfdefault}{bx}{n}
\newcommand{\tens}[1]{\bm{\mathsfit{#1}}}

\def\tR{{\tens{R}}}

\newcommand{\etens}[1]{\mathsfit{#1}}

\def\etR{{\etens{R}}}

\newcommand{\bsb}{\boldsymbol{b}}

\newcommand{\bsw}{{\boldsymbol{w}}}

\newcommand{\bsG}{{\boldsymbol{G}}}
\newcommand{\bsH}{{\boldsymbol{H}}}
\newcommand{\bsI}{{\boldsymbol{I}}}

\newcommand{\bsL}{{\boldsymbol{L}}}

\newcommand{\bsO}{{\boldsymbol{O}}}

\newcommand{\bsR}{{\boldsymbol{R}}}

\newcommand{\bsU}{{\boldsymbol{U}}}
\newcommand{\bsV}{{\boldsymbol{V}}}

\newcommand{\bsX}{{\boldsymbol{X}}}
\newcommand{\bsY}{{\boldsymbol{Y}}}

\newcommand{\bmu}{\boldsymbol{\mu}}
\newcommand{\bsigma}{\boldsymbol{\sigma}}

\newcommand{\figref}[1]{Fig.~\ref{#1}}

\newcommand{\secref}[1]{Section~\ref{#1}}

\newcommand{\tabref}[1]{Table~\ref{#1}}

\usepackage{mdframed}
\definecolor{mdproofbg}{rgb}{0.95,0.95,0.95}
{\begin{mdframed}[backgroundcolor=mdproofbg,linewidth=0]\begin{proof}}%
{\end{proof}\end{mdframed}}

\definecolor{mdworkingbg}{rgb}{1.0,0.95,0.95}
{\begin{mdframed}[backgroundcolor=mdworkingbg,linewidth=0]\begin{minipage}{\columnwidth}}%
{\end{minipage}\end{mdframed}}

\usepackage{xpatch}
\makeatletter
\AtBeginDocument{\xpatchcmd{\@thm}{\thm@headpunct{.}}{\thm@headpunct{:}}{}{}} 
\AtBeginDocument{\xpatchcmd{\@thm}{\fontseries\mddefault\upshape}{}{}{}} 
\makeatother

\definecolor{bgCode}{rgb}{0.98, 0.98, 0.98}
\definecolor{codegray}{rgb}{0.5,0.5,0.5}
\lstnewenvironment{pycode}%
{  \noindent
	\minipage{\linewidth} 
	\bigskip 
	\lstset{language=Python,numbers=left,numbersep=5pt}
	\raggedright
	\scriptsize{\textsf{Python Code}\vspace{-1.75ex}}
	\raggedleft}%
{\smallskip \endminipage}
\usepackage{url}
\usepackage{booktabs}
\usepackage{makecell}
\usepackage{multirow}
\usepackage{color}
\usepackage{colortbl}
\usepackage{threeparttable}

\usepackage{hyperref}

\usepackage[accepted]{icml2025}

\usepackage{amsmath}
\usepackage{amssymb}
\usepackage{mathtools}
\usepackage{amsthm}

\usepackage[capitalize,noabbrev]{cleveref}

\theoremstyle{plain}

\theoremstyle{definition}

\theoremstyle{remark}

\usepackage[textsize=tiny]{todonotes}

\begin{document}

\twocolumn[
\icmltitle{Can We Predict Performance of Large Models across Vision-Language Tasks?}



\icmlsetsymbol{equal}{*}

\begin{icmlauthorlist}
\icmlauthor{Qinyu Zhao}{anu}
\icmlauthor{Ming Xu}{anu}
\icmlauthor{Kartik Gupta}{sm}
\icmlauthor{Akshay Asthana}{sm}
\icmlauthor{Liang Zheng}{anu}
\icmlauthor{Stephen Gould}{anu}
\end{icmlauthorlist}

\icmlaffiliation{anu}{Australian National University, Canberra, Australia}
\icmlaffiliation{sm}{Seeing Machines Ltd, Canberra, Australia}

\icmlcorrespondingauthor{Qinyu Zhao}{qinyu.zhao@anu.edu.au}

\icmlkeywords{Machine Learning, ICML}

\vskip 0.3in
]



\printAffiliationsAndNotice{}  

\begin{abstract}
Evaluating large vision-language models (LVLMs) is very expensive, due to high computational cost and the wide variety of tasks. The good news is that if we already have some observed performance scores, we may be able to infer unknown ones. In this study, we propose a new framework for predicting unknown performance scores based on observed ones from other LVLMs or tasks. We first formulate the performance prediction as a matrix completion task. Specifically, we construct a sparse performance matrix $\boldsymbol{R}$, where each entry $R_{mn}$ represents the performance score of the $m$-th model on the $n$-th dataset. By applying probabilistic matrix factorization (PMF) with Markov chain Monte Carlo (MCMC), we can complete the performance matrix, \ie predict unknown scores. Additionally, we estimate the uncertainty of performance prediction based on MCMC. Practitioners can evaluate their models on untested tasks with higher uncertainty first, which quickly reduces the prediction errors. We further introduce several improvements to enhance PMF for scenarios with sparse observed performance scores. Our experiments demonstrate the accuracy of PMF in predicting unknown scores, the reliability of uncertainty estimates in ordering evaluations, and the effectiveness of our enhancements for handling sparse data. Our code is available at \url{https://github.com/Qinyu-Allen-Zhao/CrossPred-LVLM}.
\end{abstract}

\begin{figure}[tb]
\centering
\includegraphics[width=0.8\textwidth,trim=0 20 40 0,clip]{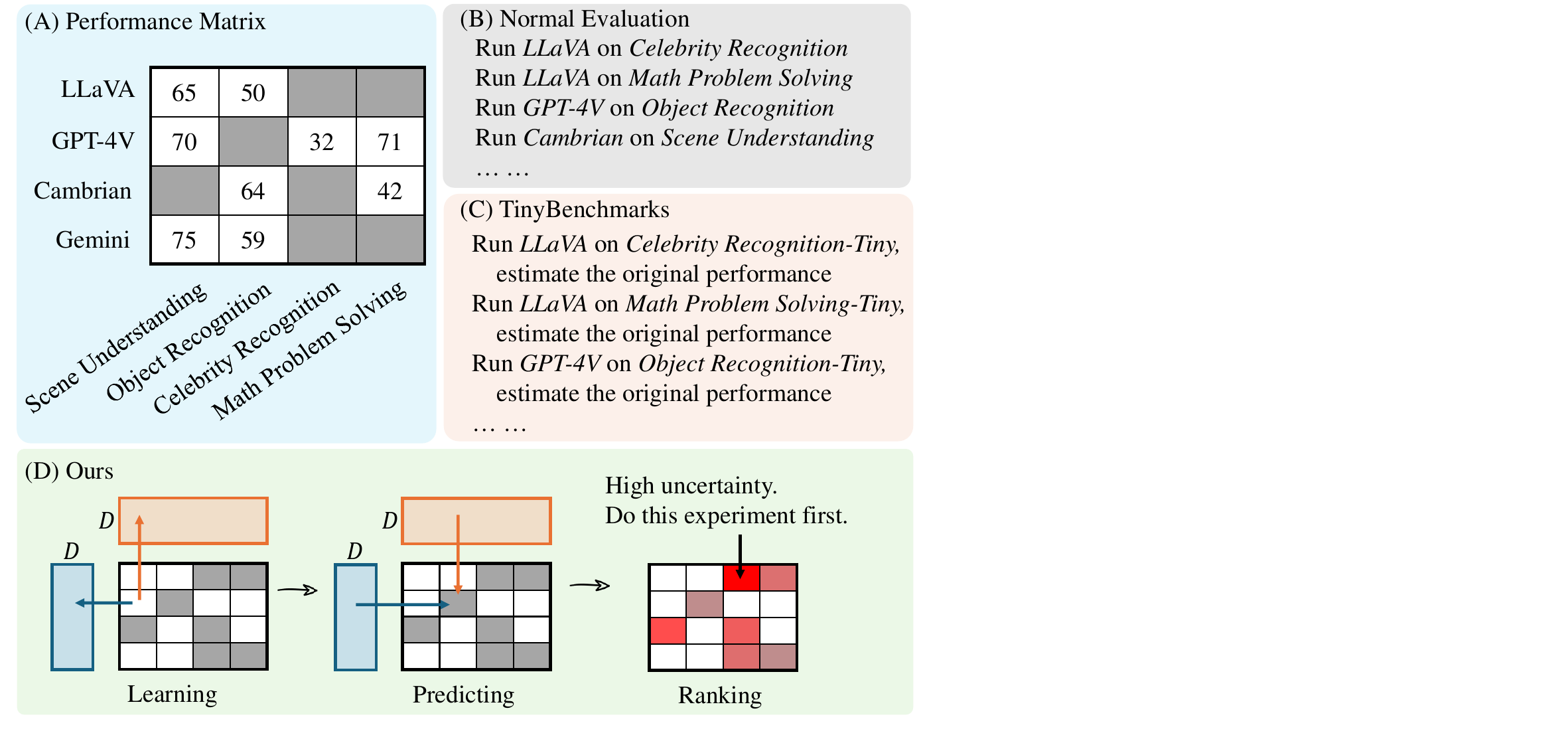}
  \caption{\textbf{Framework}. (A) Given a sparse matrix of performance scores of LVLMs on various tasks, the goal is to estimate the missing entries. (B) A normal way is to evaluate untested model-dataset pairs one-by-one. (C) TinyBenchmarks~\citep{tinybenchmarks} runs models on smaller test sets and reproduce the original performance. (D) We use PMF to predict missing entries, reducing unnecessary evaluations, and rank new experiments based on uncertainty.} 
  \label{fig:framework}
\end{figure}

\section{Introduction} \label{sec:introduction}
Large vision-language models (LVLMs) have shown remarkable performance across a wide range of multimodal tasks~\citep{llava_v1,llava_1_5,wang2024qwen2,bai2025qwen2_5,lu2024deepseek,internvl_1_5,gpt4,gemini15}. However, due to their large scale and versatility, evaluating these models leads to high cost.

First, large-scale models result in significant computational and memory costs. Second, since a single LVLM can handle various tasks, hundreds of benchmarks have been proposed to comprehensively assess the strengths and weaknesses of models~\citep{li2024survey,seed2,fu2023mme}. Last, increase in model size, the development of video LVLMs~\citep{bai2025qwen2_5}, and test-time scaling~\cite{O1} techniques further raise evaluation cost. In our estimation, evaluating a 7B model on 50 benchmarks takes around 20 hours on one A100 GPU, and even three times longer when using test-time scaling or on video benchmarks~\citep{videomme,longvideobench} (Details in \secref{asec:eval_cost}).

Fortunately, we have already observed performance scores from some of these models on some tasks, for instance, from the official reports of released models and datasets. For new models, scores can also be readily obtained with limited compute by running on a small number of tasks. If these observed scores can be used to predict unknown ones, we could avoid unnecessary evaluations and effectively reduce costs. Recent works~\citep{tinybenchmarks,lmms_eval}  require running the same model on the same task to predict model performance, and most of them ignore the potential of observed performance data from other models or tasks.

In this study, we propose a new framework for predicting unknown performance scores based on observed ones from other LVLMs or tasks. We first formulate this as a matrix completion problem. Specifically, we construct a sparse performance matrix $\bsR$ where each entry $R_{mn}$ represents the performance score of the $m$-th model on the $n$-th dataset. By applying probabilistic matrix factorization (PMF) with Markov chain Monte Carlo (MCMC), we can predict unknown performance scores based on observed entries in the matrix. A summary of the framework is shown in \figref{fig:framework}.

A bonus of our framework is active evaluation, which aims to select a subset of model-dataset pairs to evaluate in order to minimize prediction errors across the entire performance matrix. Given a PMF model on a very sparse performance matrix, we calculate prediction uncertainty from MCMC and prioritize evaluating model-dataset pairs with high uncertainty. Our experiments will confirm the effectiveness of this strategy for active evaluation.

A challenge is that PMF tends to predict the average score for models and datasets with very few observed scores, resulting in poor prediction results~\citep{pmf}. To address this, we introduce several improvements to enhance PMF for scenarios with sparse observed data. First, we extend PMF to a simple tensor factorization approach, which can handle multiple performance metrics across different vision-language tasks. Second, we utilize Bayesian PMF~\citep{bpmf} with an LKJ prior~\citep{lkj} on the variance. Third, we also incorporate specific information from LVLMs and datasets to improve performance prediction. For example, if we know a model uses CLIP as a vision encoder, the information may help predict the model's performance, especially when we observe only a few performance scores.

In experiments, we conduct a systematic evaluation of 108 LVLMs across 176 distinct datasets derived from 36 existing benchmarks, based on prior works~\citep{VLMEvalKit,lmms_eval,hemm,prismatic,universal}. We evaluate open-source models such as LLaVA-v1.5~\citep{llava_1_5}, InstructBLIP~\citep{instructblip}, mPLUG-Owl~\citep{ye2023mplug2}, and MiniGPT-4~\citep{minigpt4}, as well as closed-source models including GPT-4o, GPT-4~\citep{gpt4}, Gemini-1.5~\citep{gemini15}. The benchmarks cover general VQA~\citep{seed2}, knowledge-dense VQA~\citep{mmmu}, hallucination~\citep{pope_benchmark}, medicine~\citep{pathvqa}, emotion recognition~\citep{goodfellow2013challenges}, and others. To reduce computational and API costs, we subsample some datasets, following \citet{hemm}.

Using the results from 108 LVLMs across 176 datasets, we construct a $108 \times 176$ performance matrix, with some entries masked for testing. First, experiments demonstrate that PMF accurately predicts masked scores and consistently outperforms baselines as long as more than 10\% entries in the performance matrix are observed. Second, we also show that selecting high-uncertainty model-dataset pairs for evaluation significantly reduces prediction errors compared to random selection. third, our improvements effectively alleviate the sparse data issue of PMF. Last, we also expand the performance matrix by introducing new models and datasets, showing the generalization ability of our method.

In summary, this paper covers three main points. First, we formulate a problem of predicting the unknown performance of LVLMs across tasks. Second, we apply the well-established PMF algorithm to this problem, show the application of active evaluation, and propose several strategies to mitigate the sparse data issue. Third, we conduct a comprehensive evaluation of LVLMs, constructing training and testing sets for further experiments.

\section{Related Works} \label{sec:related}
\subsection{Benchmarking LVLMs}
In recent years, an increasing number of LVLMs demonstrate impressive capabilities on a wide variety of tasks, including GPT-4~\citep{gpt4}, Gemini~\citep{team2023gemini}, LLaVA~\citep{llava_v1,llava_1_5}, InstructBLIP~\citep{instructblip},  InternVL~\citep{internvl,internvl_1_5}, and Prismatic VLMs~\citep{prismatic}. Their versatility requires various benchmarks to fully understand their strengths and weaknesses. Existing datasets can be repurposed for assessing these models, such as Flickr30k~\citep{flickr30k}, GQA~\citep{gqa}, and OKVQA~\citep{okvqa}. Recent works also propose new benchmarks to evaluate LVLMs in handling complex reasoning tasks, including SEED-Bench-2~\citep{seed2}, MMMU~\citep{mmmu}, and MME~\citep{fu2023mme}. Additionally, as LVLMs become more integrated into everyday applications, benchmarks~\citep{pope_benchmark} have been introduced to assess trustworthy issues like hallucination in these models. The variety of LVLMs and benchmarks leads to substantial computational demands and memory usage.

\subsection{Improve Evaluation Efficiency}
Recent works introduce unified frameworks to assess models across multiple benchmarks using a single codebase, such as VLMEvalKit~\citep{VLMEvalKit}, LMMs-Eval~\citep{lmms_eval}, and HEMM~\citep{hemm}. Our study builds on these efforts by consolidating their frameworks and integrating models in Prismatic VLMs series~\citep{prismatic,universal}.

Predicting unknown model performance can reduce the evaluation cost. Prior works select a coreset of samples from a large benchmark, for evaluating large language models (LLMs)~\citep{tinybenchmarks,efficient_benchmarking} and LVLMs~\citep{lmms_eval,lime}. The performance of a specific model on the coreset is used to estimate its performance on the full benchmark. Besides, existing studies estimate model performance on an unlabeled test set based on distribution shift~\citep{autoeval}, confidence scores~\citep{doc,bbs}, or LLM feedback~\citep{llm_as_judge}. Instead of running models on a coreset or an unlabeled set, our framework predicts unknown performance by utilizing the correlation between model performances across benchmarks, and is complementary to coreset-based ones (see \secref{sec:coreset_plus_ours}).

A concurrent study by~\citet{zhang2024collaborative} applies deep matrix factorization to predict LLM performance. Their goal and methodology differ from ours. First, their work aims to explore whether LLM performance can be predicted except using scaling laws, while we focus on improving evaluation efficiency. Second, they use neural networks for prediction, whereas we adopt PMF with MCMC, which allows us to quantify uncertainty in performance prediction.

Another related direction is adaptive testing~\citep{irt_eval,lifelong}, \ie adaptively selecting a subset of samples for evaluation. While prior work relies on statistically inferred sample difficulty, we propose a method to rank model-dataset pairs for evaluation based on uncertainty in performance prediction from MCMC.

\subsection{Probabilistic Matrix Factorization}
PMF~\citep{pmf} is widely applied in recommender systems. Given part of user ratings for items, the goal is to model the observed ratings and predict the missing ones. PMF achieves this by decomposing the observed rating matrix into two lower-dimensional matrices, representing the latent features of users and items. A rating is modeled via a Gaussian distribution centered around the dot product of the user’s and item’s feature vectors.

One major challenge with PMF is that, if users rate very few items, their predicted ratings will be near the average for those items. Bayesian PMF (BPMF)~\citep{bpmf} addresses this by placing distributions over the priors of the latent user and item features, making it more effective in handling sparse data. Additionally, Constrained PMF~\citep{pmf} introduces a latent similarity constraint matrix to address the problem.

\section{Modeling LVLM Performance}
In this section, we first describe the application of PMF to model the performance score matrix of LVLMs across datasets. Then, we discuss active evaluation for LVLMs. Last, three techniques are introduced to enhance PMF: supporting multiple metrics, incorporating Bayesian PMF, and integrating model and dataset profiles in modeling.

\subsection{Revisit PMF}
Let $\bsR$ be an $M \times N$ matrix representing model performance scores on datasets, where $M$ is the number of models and $N$ is the number of datasets. For simplicity, we initially assume a single performance metric. In reality, benchmarks often employ multiple metrics, so $\bsR$ becomes an $M \times N \times S$ tensor, where $S$ represents the total number of metrics. We will extend our framework to this in \secref{sec:multiple_metrics}.

In practice, only a subset of the elements in $\bsR$ are observed, meaning we evaluate only a portion of the model-dataset pairs and aim to estimate the remaining performance scores. Specifically, we define a matrix $\bsO \in \{0,1\}^{M \times N}$, where $O_{mn} = 1$ if $R_{mn}$ is observed, and $0$ otherwise.

\begin{figure}
\centering
\includegraphics[width=0.49\textwidth,trim=0 160 100 5,clip]{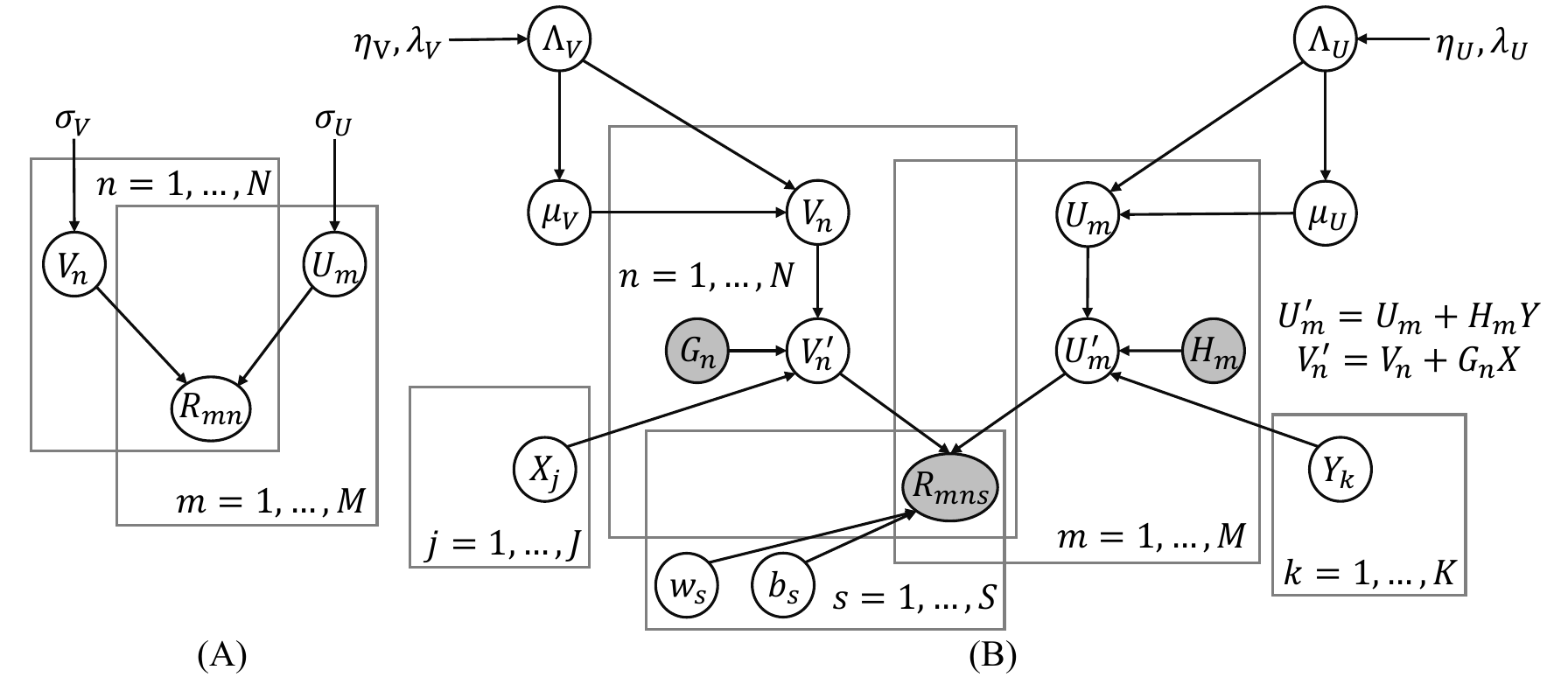}
  \caption{\textbf{Graphical Models} of PMF (A) and the enhanced model (B). (A) is adapted from the original paper~\citep{pmf}. In (B), we set the mean to $\boldsymbol{0}$ and the covariance to the identity matrix, thus omitting most of the hyper-parameters for the random variable distributions.} 
  \label{fig:graphical_model}
\end{figure}

To model the observed matrix and estimate the unknown values, we employ PMF~\citep{pmf}, as illustrated by the probabilistic graphical model in \figref{fig:graphical_model}(A). PMF decomposes $\bsR$ into two low-dimensional matrices, $\bsU \in \mathbb{R}^{M \times D}$ and $\bsV \in \mathbb{R}^{N \times D}$, where $D$ is the latent dimension. Here, $\bsU_{m,:}$ and $\bsV_{n,:}$ are the latent feature vectors for the $m$-th model and the $n$-th dataset, respectively, and we refer to them as $\bsU_{m}$ and $\bsV_{n}$. These latent vectors are modeled as multivariate Gaussian distributions, and the observed ratings are assumed to follow a Gaussian distribution centered at the dot product of the latent feature vectors:
\begin{align}
    & p(\bsR \mid \bsU, \bsV, \sigma^2) = \nonumber \\
    & \prod_{m=1}^{M} \prod_{n=1}^{N} \left[ \mathcal{N} \left( R_{mn} \mid \bsU_{m}^T \bsV_{n}, \sigma^2 \right) \right]^{O_{mn}}, \\
    & p(\bsU \mid \sigma_U^2) = \prod_{m=1}^{M} \mathcal{N}(\bsU_{m} \mid \mathbf{0}_D, \sigma_U^2 \bsI_D), \\
    & p(\bsV \mid \sigma_V^2) = \prod_{n=1}^{N} \mathcal{N}(\bsV_{n} \mid \mathbf{0}_D, \sigma_V^2 \bsI_D),
\end{align}
where $\bsI_D$ is a $D \times D$ identity matrix, and $\mathcal{N}(x \mid \mu, \sigma^2)$ represents the probability density function of a Gaussian distribution with mean $\mu$ and variance $\sigma^2$. For simplicity, we set $\sigma_U=\sigma_V=1$.

Rather than using Maximum A Posteriori estimation to obtain point estimates of the unknown performance scores in $\bsR$, we apply MCMC to obtain distributions over the estimated scores and quantify prediction uncertainty.

We compare PMF and deep learning-based matrix completion methods, such as Deep Matrix Factorization (DMF)~\citep{dmf} and Graph Convolutional Matrix Completion (GCMC)~\citep{gcmc}. The results indicate that PMF is more suitable to our problem than deep learning methods, probably because deep learning is more data hungry. Thus, this study will focus on PMF. The details are shown in the supplementary materials (\secref{sec:app_comp_pmf_dmf}).

\subsection{Active Evaluation}
MCMC allows us to estimate score distributions and readily obtain uncertainty estimates for each unknown score, enabling us to prioritize evaluation experiments. For example, if we are uncertain about GPT-4's performance on 3D understanding but confident about LLaVA's performance on object recognition, we can prioritize evaluating GPT-4 on the 3D task when our resources are limited.

In our method, we begin by applying PMF to model a sparse performance matrix. Using MCMC, we get hundreds of estimations of each unknown score and calculate the standard deviation of estimations as a measure of uncertainty. The unobserved scores are ranked by their uncertainties. High-uncertainty scores are replaced with ground truth, simulating evaluation process in practice. We rerun PMF with updated observed data, calculate uncertainty, and determine the next set of evaluations. This process is repeated until our resource budget is exhausted or all scores are observed.

\subsection{Multiple Metrics}\label{sec:multiple_metrics}
Our experiments show that standard PMF performs well with sufficient observed data. But its performance degrades significantly and is even worse than predicting the average score, when the observed data is very sparse (\ie fewer than 10\% model-dataset pairs are observed). To address this, we enhance our model with several techniques, with a new graphical model shown in \figref{fig:graphical_model}(B).

Previously, we assumed that each dataset has only one scoring metric, but this is usually not the case in practice. For example, yes-or-no questions can be evaluated using accuracy, precision, recall, and F1 score, while open-ended questions may use metrics like BART score~\citep{bartscore} and BERT score~\citep{bertscore}. Model performances are represented by a tensor $\tR \in \mathbb{R}^{M \times N \times S}$, where $S$ is the total number of metrics. Empirically, we find that using PMF to model and predict each metric independently works well when sufficient data is available. However, when observed data is sparse, incorporating relationships between metrics will be helpful to performance prediction.

To address this, we extend our PMF model into a simple Probabilistic Tensor Factorization (PTF), where we decompose the 3D tensor $\tR$ into the product of two low-rank matrices and a 1D vector. This can be interpreted as applying a linear transformation to the original PMF output, translating it into multiple metrics. Specifically, we define:
\begin{align}
    & p(\tR \mid \bsU, \bsV, \bsw, \bsb, \sigma^2) = \nonumber \\
    & \prod_{m=1}^{M} \prod_{n=1}^{N} \prod_{s=1}^{S} \left[ \mathcal{N} \left( \etR_{mns} \mid (\bsU_{m}^T \bsV_{n})w_s + b_s, \sigma^2 \right) \right]^{O_{mns}}, \\
    & p(\bsw \mid \sigma_w^2) = \mathcal{N}(\bsw \mid \mathbf{0}_S, \sigma_w^2 \bsI_S), \\
    & p(\bsb \mid \sigma_b^2) = \mathcal{N}(\bsb \mid \mathbf{0}_S, \sigma_b^2 \bsI_S),
\end{align}
where we set $\sigma_w = \sigma_b = 1$ for simplicity.

This approach implicitly assumes a linear relationship between scoring metrics, which may not exactly hold in reality. However, we usually observe some linear correlation between the metrics on the same task. More sophisticated techniques, such as advanced tensor factorization methods are left as future work to further improve the model.

Note that some metrics may be irrelevant for certain datasets, e.g., accuracy is not meaningful for long-answer questions. While our model can predict these scores, we discard the predicted results because there is no ground truth.

\subsection{Bayesian PMF}
Instead of using fixed priors for the feature vectors, we model the priors using probabilistic distributions, as proposed by \citet{bpmf}. Unlike the original paper, which employs a Wishart distribution for the variance, we use the LKJ correlation prior~\citep{lkj} and an Exponential prior to model the variance, as suggested by the PyMC official documentation,
\begin{equation}
    \boldsymbol{\Lambda}^{-1}_U = (\diag{\bsigma_L}\bsL_U)(\diag{\bsigma_L}\bsL_U)^T,
\end{equation}
where $p(\bsL_U \mid \eta_U) = \text{LKJ}(\bsL_U\bsL_U^T \mid \eta_U)$ and $p(\bsigma_L \mid \lambda_U) = \prod_{d=1}^{D}\text{Exp}(\sigma_d \mid\lambda_U)$. Then, latent feature vectors are modeled as follows:
\begin{align}
    & p(\bmu_U \mid \boldsymbol{\Lambda}^{-1}_U) = \mathcal{N}(\bmu_U \mid \mathbf{0}_D, \boldsymbol{\Lambda}^{-1}_U), \\
    & p(\bsU \mid \bmu_U,\boldsymbol{\Lambda}^{-1}_U) = \prod_{m=1}^{M} \mathcal{N}(\bsU_{m} \mid \bmu_U, \boldsymbol{\Lambda}^{-1}_U).
\end{align}
A similar formulation applies to $\bsV$, which we omit here.

\begin{figure*}
\centering
\includegraphics[width=0.98\textwidth,trim=10 170 0 0,clip]{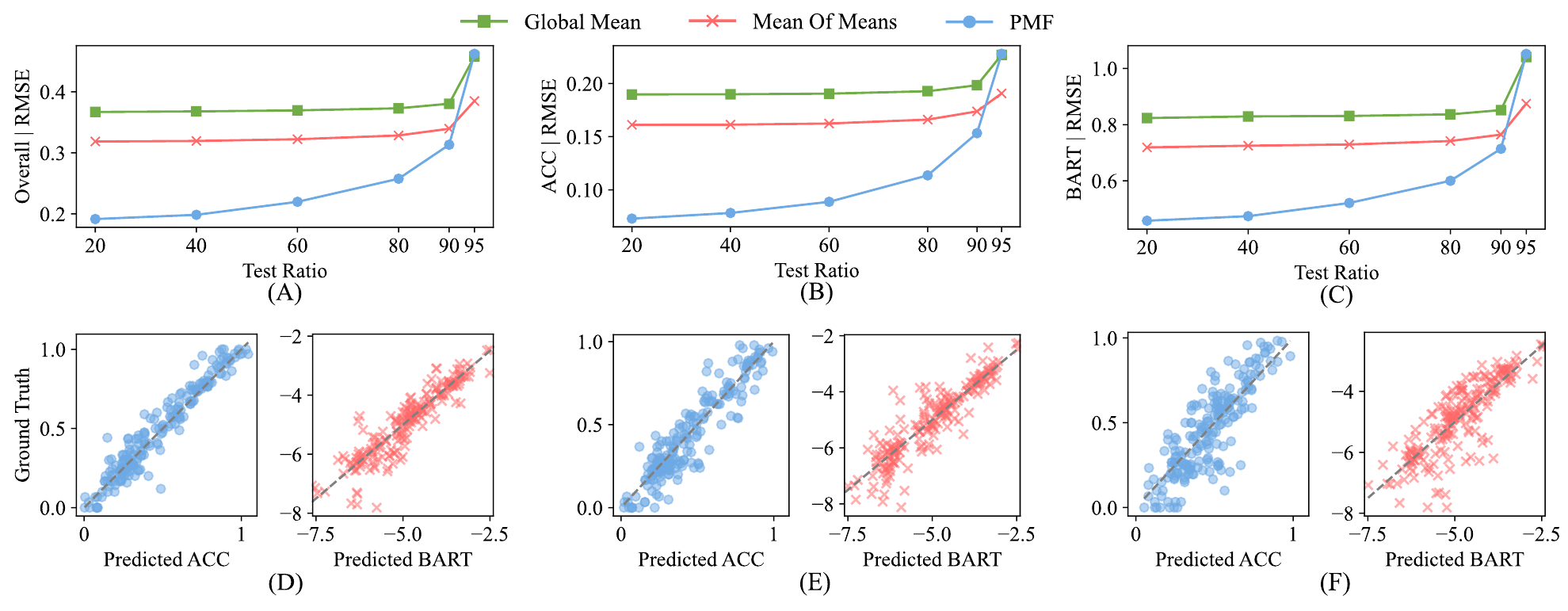}
  \caption{\textbf{Performance of PMF.} (A-C) PMF consistently outperforms both baselines when the test ratio is below 90\% for estimating all unobserved scores (A), accuracy scores (B), and BART scores (C), with particularly strong performance at lower test ratios. (D-F) The predicted scores exhibit correlations with the ground truth at test ratios of 20\% (D), 60\% (E), and 90\% (F). Gray dashed lines represent perfect prediction \ie $y=x$. We subsampled 200 scores in (D-F) for visualization.} 
  \label{fig:result_of_pmf}
\end{figure*}

\subsection{Model and Dataset Profiles}
The final enhancement is the incorporation of additional information about the models and datasets. For example, knowing that two LVLMs use CLIP as the vision encoder, or that LLaVA-v1.5 and LLaVA-NeXT are developed by the same team, suggests potential relationships in their performances. Inspired by Constrained PMF~\citep{pmf}, we incorporate extra information as model and dataset profiles, to improve performance prediction.

Let $\bsH \in \mathbb{R}^{M \times K}$ and $\bsG \in \mathbb{R}^{N \times J}$ represent the model and dataset profiles, where $\bsH_{m,:}$ encodes $K$ properties of the $m$-th model (e.g., vision encoder type), and $\bsG_{n,:}$ encodes $J$ properties of the $n$-th dataset. We introduce Gaussian-distributed variables $\bsY \in \mathbb{R}^{K \times D}$ and $\bsX \in \mathbb{R}^{J \times D}$ to learn the effects of these profiles. The latent feature vectors are now the sum of the original vectors and the profile features, following Constrained PMF~\citep{pmf}.
\begin{align}
    & p(\bsY \mid \sigma_Y^2) = \prod_{k=1}^{K} \mathcal{N}(\bsY_{k} \mid \mathbf{0}_D, \sigma_Y^2 \bsI_D),\\
    & p(\bsX \mid \sigma_X^2) = \prod_{j=1}^{J} \mathcal{N}(\bsX_{j} \mid \mathbf{0}_D, \sigma_X^2 \bsI_D), \\
    & \bsU' = \bsU + \bsH \bsY, \quad \bsV' = \bsV + \bsG \bsX.
\end{align}

\textbf{Oracle Profiles.} To explore the upper bound of model and dataset similarities, we use the full $\bsR$ matrix to cluster models and datasets. For each model, we take $\bsR_{i,:}$ (its performance across all datasets) as a vector and apply the K-Means algorithm to cluster all models. We select the optimal number of clusters using the elbow method. Similarly, for each dataset, we cluster $\bsR_{:,j}$ in the same way. We convert the cluster assignments into one-hot vectors as profiles.

\textbf{Custom Profiles.} Since oracle profiles rely on complete performance data, they are not feasible for real-world use. To overcome this, we define custom profiles that can be applied in practice. For models, we include features such as the number of parameters in the LLM backbone, vision encoder type (one-hot), and the LVLM family (one-hot), illustrated in the supplementary material (\tabref{tab:model_info}). Additionally, we cluster datasets based on latent representations obtained from various models and get one-hot encoded dataset profiles. We explore three different approaches to generate these latent representations: D1. using MPNet~\citep{mpnet} to encode a short description of each dataset. D2. using CLIP to encode images and BGE-M3 to encode questions in a dataset (following~\citet{lmms_eval}), then averaging the embeddings on the dataset; and D3. using LLaVA-7B to encode both images and text, then averaging the embeddings as the profile for the dataset.

\section{Experiments}
We first construct a large performance matrix by evaluating LVLMs on different benchmarks, and then, we present key experiments to validate our framework.

\subsection{Evaluating Models on Benchmarks}
Based on prior code repositories~\citep{VLMEvalKit,lmms_eval,hemm}, we evaluate 108 LVLMs on 176 tasks in 36 benchmarks and build a $108\times 176$ performance matrix, which is used for our main experiments on PMF and MCMC in \secref{sec:est} and \ref{sec:act_eval}.

The open-source models we cover include LLaVA-v1.5~\citep{llava_1_5}, LLaVA-NeXT~\citep{llava_1_5}, InstructBLIP~\citep{instructblip}, mPLUG-Owl~\citep{ye2023mplug2}, and Prismatic VLMs~\citep{prismatic}. We also evaluate closed-source models like GPT-4~\citep{gpt4} and Gemini-1.5~\citep{gemini15}. 

The benchmarks span a variety of domains, including general VQA (SEED-2), knowledge-dense VQA (MMMU), hallucination (POPE), medical question answering (PathVQA), and emotion recognition (FaceEmotion). Some large-scale benchmarks, such as SEED-2~\citep{seed2} and MMMU~\citep{mmmu}, cover multiple tasks. To conduct a fine-grain analysis, we split these benchmarks into task-specific datasets, resulting in 176 datasets in total. Following HEMM~\citep{hemm}, we subsample some datasets to reduce computational and API calling costs of LVLMs. For each dataset, we calculate a main metric for PMF (either accuracy or BARTScore), and several other metrics, leading to a total of six metrics for PTF modeling. 

To evaluate the generalization ability of our framework, we expand the model and dataset pool by introducing \textbf{new models}: Qwen2-VL-Instruct 2B / 7B~\citep{wang2024qwen2}, Qwen2.5-VL-Instruct 3B / 7B / 32B~\citep{bai2025qwen2_5}, and DeepSeek-VL tiny / small~\citep{lu2024deepseek}, as well as \textbf{new benchmarks}: MathVision~\citep{mathvision}, EMMA~\citep{emma}, Video-MME~\citep{videomme}, and LongVideoBench~\citep{longvideobench}. This expansion increases the performance matrix to cover 115 LVLMs across 40 benchmarks. We present the results in \secref{sec:gen_to_new}.

Full details of datasets and models are provided in the supplementary material (\secref{asec:evaluating_details}).

\subsection{Estimating Unknown Performance}\label{sec:est}
We mask $P\%$ of the elements in the score matrix $\bsR$, use the observed portion to normalize $\bsR$, and train the PMF model using MCMC sampling. The model reconstructs the matrix $\hat{\bsR}$, and we evaluate the performance by comparing the estimated values with the ground truth for the masked elements. For MCMC, we use the No-U-Turn Sampler (NUTS)~\citep{nuts}, an advanced Hamiltonian Monte Carlo method~\citep{hmc}, tuning with 500 samples in the burn-in stage and drawing 100 samples. Empirical results show that 100 samples are sufficient for stable estimation. The mean prediction from MCMC is the final reconstructed matrix $\hat{\bsR}$.

We use Root Mean Squared Error (RMSE) as the primary metric. Additional metrics such as Mean Absolute Error (MAE) and the coefficient of determination ($R^2$), as well as ranking-based metrics including Spearman’s rank correlation, Kendall rank correlation, Precision and Recall are reported in the supplementary material (\secref{sec:more_results}).

We compare our method against two baselines: (1) Global Mean: predicting the global mean for unobserved scores; (2) Mean of Means: for each unobserved score, we average the mean performance of the model, the mean performance on the dataset, and the global mean.

\textbf{Results.} As shown in \figref{fig:result_of_pmf}(A-C), PMF significantly outperforms the baselines when the test ratio is lower than 90\%. This suggests that when only a portion of the scores is available, PMF can infer the unobserved scores with high accuracy. Additionally, as demonstrated in \figref{fig:result_of_pmf}(D-F), the estimated scores strongly correlate with the actual scores, indicating reliable predictions from our framework.

However, as the amount of observed data decreases, PMF's performance declines as can be expected. In extreme cases where the test ratio exceeds 90\%, with limited information about model or dataset performance, PMF can perform worse than predicting the means.

\begin{figure}[tb]
\centering
\includegraphics[width=0.4\textwidth,trim=0 0 0 10,clip]{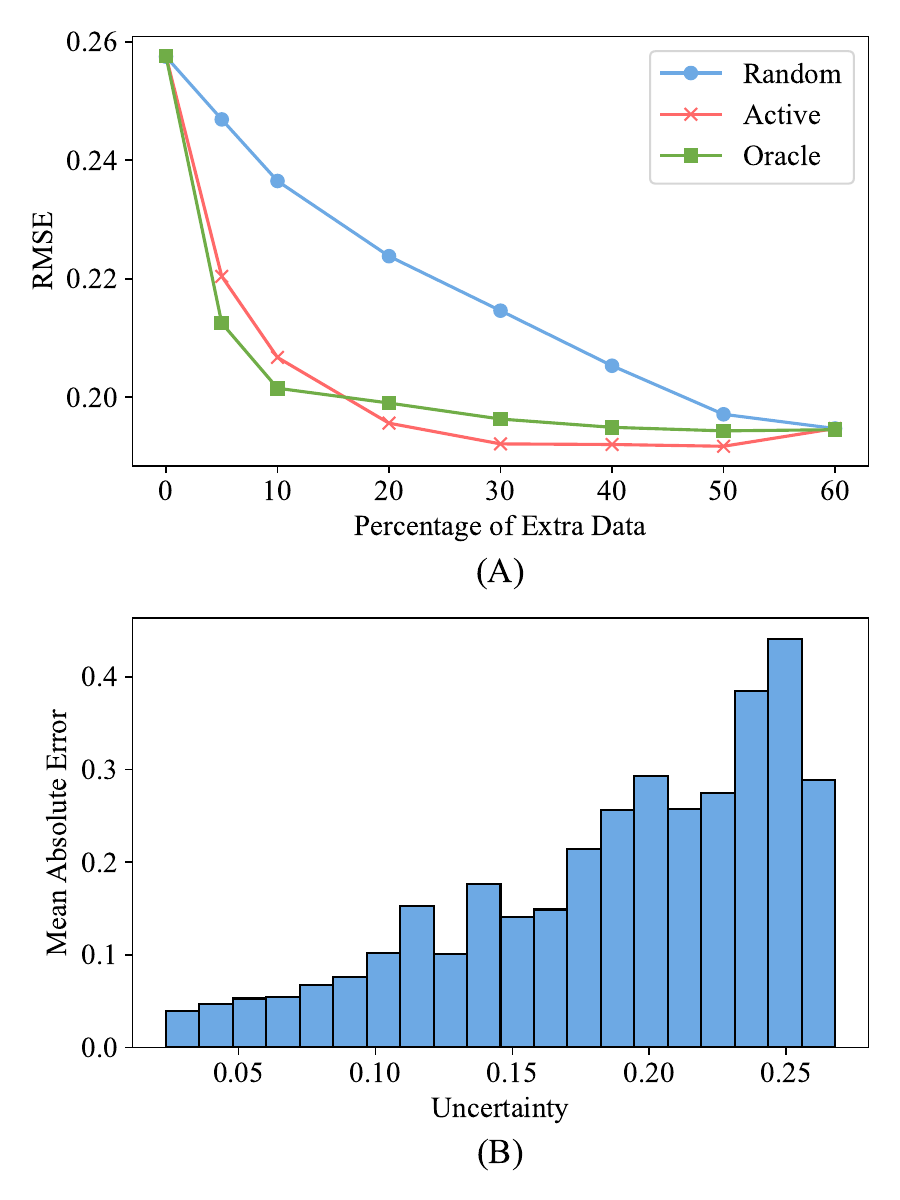}
  \caption{\textbf{Comparison of Active Evaluation Methods.} Starting with 20\% of the data observed, we progressively conduct additional LVLM evaluations using three different strategies. (A) RMSE improvement demonstrate the advantage of our method compared to random evaluation. (B) Uncertainties from MCMC are correlated with the actual absolute errors.}
  \label{fig:result_of_active_eval}
\end{figure}

\begin{table}[tb]
\scriptsize 
  \centering
  \caption{\textbf{Comparison of PMF and PTF.} Superior results are highlighted. PMF (Sep) models each score separately, while PMF (OneMat) combines accuracy and BART scores into a single matrix, as each dataset contains either accuracy or BART scores. PTF is the enhanced model that supports multiple scoring metrics, which outperforms PMF at a high test ratio.}
\setlength{\tabcolsep}{1.2mm}{
    \begin{tabular}{lcccccccc}
    \toprule
    Method & Overall & Acc & Precision & Recall & F1 & BART & BERT \\ 
    \midrule
    \multicolumn{2}{l}{\textit{Test Ratio: 20\%}} \\
    PMF (Sep) & \cellcolor{orange!20}0.175 & \cellcolor{orange!20}0.073 & 0.135 & \cellcolor{orange!20}0.166 & 0.134 & 0.463 & \cellcolor{orange!20}0.068 \\
PMF (OneMat) & 0.193 & 0.074 & - & - & - & \cellcolor{orange!20}0.461 & - \\
PTF & 0.205 & 0.078 & \cellcolor{orange!20}0.129 & 0.176 & \cellcolor{orange!20}0.108 & 0.563 & 0.077 \\
\cmidrule{1-8}
    \multicolumn{2}{l}{\textit{Test Ratio: 90\%}} \\
    PMF (Sep) & 0.327 & 0.159 & 0.238 & 0.262 & 0.227 & 0.864 & 0.096 \\
PMF (OneMat) & 0.317 & \cellcolor{orange!20}0.156 & - & - & - & \cellcolor{orange!20}0.723 & - \\
PTF & \cellcolor{orange!20}0.290 & 0.159 & \cellcolor{orange!20}0.186 & \cellcolor{orange!20}0.230 & \cellcolor{orange!20}0.180 & 0.754 & \cellcolor{orange!20}0.094 \\
    \bottomrule
\end{tabular}}
  \label{tab:result_of_multi_scores}%
\end{table}

\begin{figure*}[tb]
\centering
\includegraphics[width=0.98\textwidth,trim=0 30 0 0,clip]{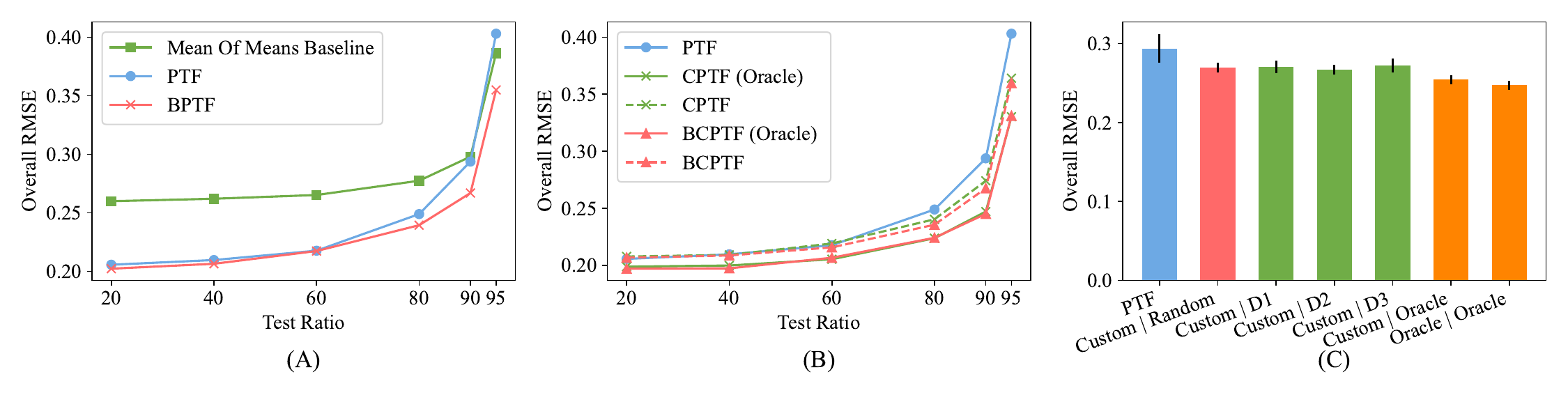}
  \caption{\textbf{Performance of Enhanced PTF.} (A) BPTF shows minimal improvement over standard PTF when data is sufficient but proves particularly beneficial under sparse conditions. (B) Custom profiles improve performance when data is limited, though a gap remains compared to oracle profiles. (C) Ablation study on model and dataset profiles. ``A $\mid$ B'' represents using A for the model profile and B for the dataset profile. Custom model profiles lead to significant performance gains, while dataset profiles contribute only marginally. BPTF, Bayesian PTF; CPTF, Constrained PTF.} 
  \label{fig:result_of_enhanced_PTF}
\end{figure*}

\subsection{Active Evaluation for LVLMs}\label{sec:act_eval}
We compare our uncertainty-based approach against two baselines: (1) random selection of model-dataset pairs, and (2) an oracle approach that selects the pairs with the highest actual errors. In the experiment, we start by 20\% data for initial training, 60\% as the pool set, and 20\% for testing.  Then, we progressively include more data from the pool set for training using three different strategies, and calculate the improvement in performance prediction on the test data. The experiment is repeated with 10 different random seeds, and we report the average improvement.

\textbf{Results.} As shown in \figref{fig:result_of_active_eval}, our uncertainty-aware method consistently outperforms the random baseline for a fixed budget of evaluations, especially when the amount of extra data is lower than 30\%. Besides, uncertainties from MCMC are correlated with the actual absolute errors.

\subsection{Enhancing PMF}
We apply three enhancement techniques to our PMF model and evaluate their effectiveness across different test ratios. To minimize experimental variance, we perform each experiment 10 times with different random seeds and report the average performance at each test ratio.

\textbf{Results.} As seen in \tabref{tab:result_of_multi_scores}, the multi-score method PTF can get better performance when the matrix is very sparse. When there is enough data, separately modeling PMF with each score works very well and is comparable to PTF. For BART and BERT scores, PMF even outperforms PTF. This is likely because PTF assumes a linear relationship between scores. When this assumption does not hold, such as in the case of BART and BERT scores, it degrades performance.

\begin{figure*}
\centering
\includegraphics[width=0.98\textwidth,trim=0 10 0 10,clip]{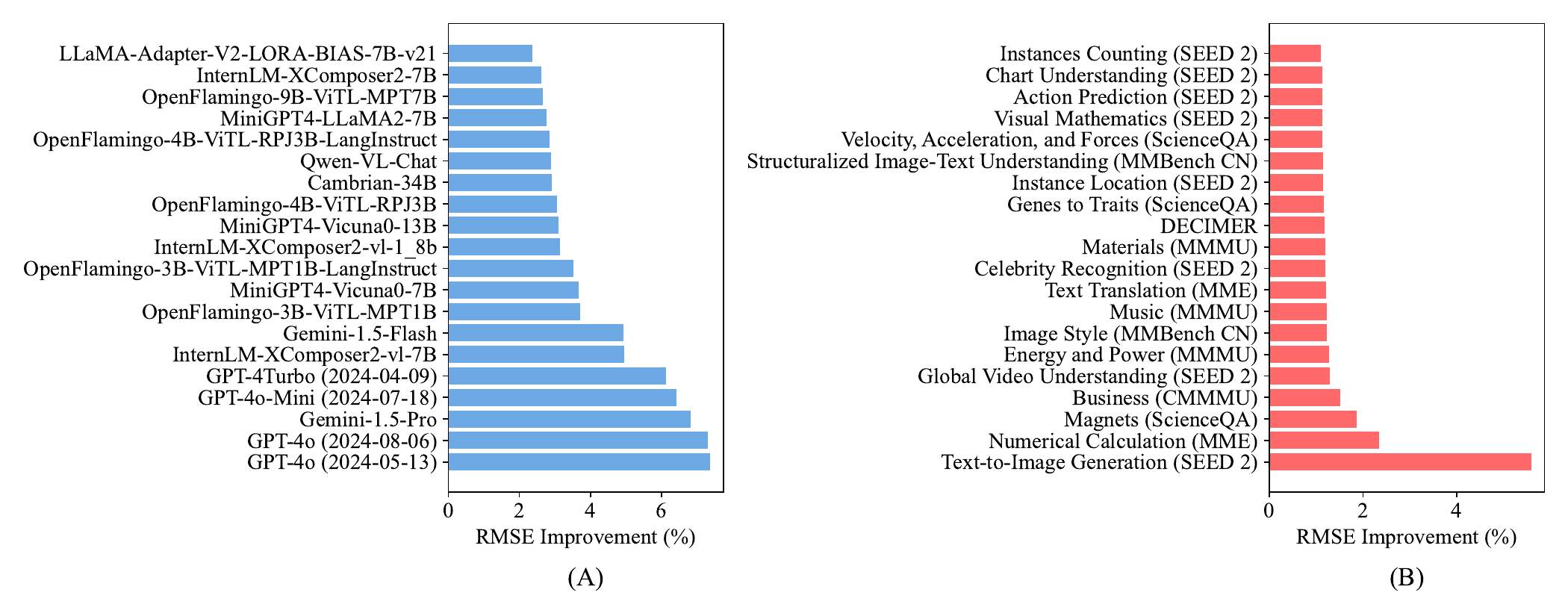}
  \caption{\textbf{Which Models and Datasets Are Informative for Performance Estimation.} Given a PMF model train on 20\% data of the performance matrix, We measure the improvement in RMSE of PMF when adding the entire results of a model (A) or a dataset (B).} 
  \label{fig:insight_informative}
\end{figure*}

\begin{table}[htbp]
\scriptsize
\centering
\caption{\textbf{Generalization to New Models and Datasets.} Average RMSEs of different methods are reported.}
\setlength{\tabcolsep}{2.0mm}{
\begin{tabular}{lccc}
\toprule
Method & New Model & New Dataset & Both New \\
\midrule
\multicolumn{4}{l}{\textit{If we only know 20\% performance of new models and new datasets}} \\
Global Mean     & 0.390 & 0.043 & 0.106 \\
Mean of Means   & 0.303 & 0.037 & 0.084 \\
PMF             & 0.326 & \cellcolor{orange!20}0.032 & 0.047 \\
BCPMF           & \cellcolor{orange!20}0.297 & 0.033 & \cellcolor{orange!20}0.039 \\
\midrule
\multicolumn{4}{l}{\textit{If we know 50\% performance of new models and new datasets}} \\
Global Mean     & 0.389 & 0.045 & 0.090 \\
Mean of Means   & 0.311 & 0.039 & 0.073 \\
PMF             & 0.265 & 0.031 & \cellcolor{orange!20}0.034 \\
BCPMF           & \cellcolor{orange!20}0.228 & \cellcolor{orange!20}0.030 & \cellcolor{orange!20}0.034 \\
\bottomrule
\end{tabular}}
\label{tab:gen_to_new}
\end{table}

\figref{fig:result_of_enhanced_PTF} illustrates the impact of the other two enhancement techniques. As shown, Bayesian PTF offers only negligible improvements over standard PTF when there is enough observed data, but it is particularly beneficial in sparse conditions. In \figref{fig:result_of_enhanced_PTF}(B), our custom profiles also show improvements when data is limited, though there remains a gap between our custom profiles and the oracle profiles. Additionally, \figref{fig:result_of_enhanced_PTF}(C) highlights that adding profiles not only enhances PTF’s overall performance but also reduces instability, as seen by smaller error bars. Model profiles show significant performance gains, whereas dataset profiles contribute only marginally. Better methods for encoding and utilizing dataset information need further exploration.

\subsection{Generalization to New Models and Datasets}\label{sec:gen_to_new}
To validate the generalization ability of the framework, we expand the performance matrix by adding new models and datasets. Then, we provide the complete performance data on old models and datasets, along with partial observations for: (1) old models on new datasets, (2) new models on old datasets, and (3) new models on new datasets. Based on this setup, we predict the missing performance scores and calculate the average RMSEs for different methods.

As shown in \tabref{tab:gen_to_new}, our method shows better generalization compared to the baselines. We observe that generalizing to new datasets is relatively easier than to new models. This is probably because new datasets are often very challenging, leading to generally lower model performance and smaller RMSEs. In contrast, new models often have novel designs and remarkable improvements. This makes generalization to unseen models more difficult and needs further exploration.

\section{Discussion}
\subsection{Predictability of Model Performance}
There are two possible perspectives to understand why the model performance can be well predicted. 

\textbf{Benchmarks.} There is redundancy in existing benchmarks, as also reported by \citet{zhang2025redundancy}. Multiple benchmarks may test similar skills like math. Besides, there might be correlation across different tasks. For instance, \citet{llava_1_5} reports that an LVLM with a stronger LLM can achieve consistently better performance on different tasks.

\textbf{Models.} Similarity in model architectures and training may also contribute to correlation between performance. First, if models use the same vision encoder, e.g., CLIP, they may share similar failure cases~\citep{tong2024eyes}. Besides, many models use training data generated from GPT-4V, possibly resulting in similar strengths and weaknesses.

Predictability is valuable during model development. For example, when exploring optimal video LLM design, \citet{zohar2024apollo} evaluate their models on a reduced set of benchmarks to guide design decisions. Only at the final stage, they assess the model on the full benchmarks to report actual performance. In such cases, they could use our method to reduce unnecessary evaluation when developing models.

\subsection{Modeling Varying Settings of LVLM Evaluation.}\label{sec:modeling_settings}
Our framework is also extensible to different evaluation settings, such as various prompts or decoding strategies. There are two possible methods. \textbf{Additional models:} treat a model under different evaluation settings as different models, such as ``LLaVA (Chain-of-Thought)'' and ``LLaVA (Beam Search)''. \textbf{Additional profile:} encode evaluation settings as extra profiles into PMF.

Specifically, we evaluate LLaVA-v1.5-7B on the 27 tasks in SEED-2, with the following various evaluation settings.

\textbf{Image input.} (1) Default: use the clean images, or (2) add Gaussian noise into the images.

\textbf{Prompt.} (1) Default: prompt the model to choose option (``Answer with the option's letter from the given choices directly.''), (2) provide no hint, or (3) use the Chain-of-Thought (CoT) prompt (``Let's think step by step, and then answer with the option's letter.'').

\textbf{Model decoding.} (1) Default: greedy decoding, (2) sampling with temperature = 0.2, (3) sampling with temperature = 0.5, or (4) beam search with temperature = 0.2 and the number of beams = 10.

We add the results under different evaluation settings into our framework and simply use PMF for prediction. \tabref{tab:modeling_settings} reports RMSE of different methods and indicate that our framework can handle different evaluation settings.

\begin{table}[tb]
\scriptsize
\centering
\caption{\textbf{Prediction Accuracy under Different Evaluating Settings.} Average RMSEs of different methods are reported.}
\setlength{\tabcolsep}{0.7mm}{
\begin{tabular}{lccccccc}
\hline
\multirow{2}{*}{Method} & \multirow{2}{*}{Default} & Input & \multicolumn{2}{c}{Prompt} & \multicolumn{3}{c}{Decoding} \\
\cline{3-8} \\[-1.6ex]
& & \makecell{Gaussian\\ Noise} & No Hint & CoT & \makecell{Sampling\\ (t=0.2)} & \makecell{Sampling\\ (t=0.5)} & Beam \\
\hline
\multicolumn{4}{l}{\textit{Test Ratio: 20\%}} \\
Global Mean     & 0.112 & 0.105 & 0.090 & 0.117 & 0.127 & 0.109 & 0.111 \\
Mean of Means   & 0.090 & 0.088 & 0.090 & 0.102 & 0.105 & 0.092 & 0.088 \\
Ours (Profiles) & \cellcolor{orange!20}0.041 & 0.055 & 0.075 & 0.064 & 0.045 & 0.055 & 0.052 \\
Ours (Models)   & 0.043 & \cellcolor{orange!20}0.045 & \cellcolor{orange!20}0.073 & \cellcolor{orange!20}0.050 & \cellcolor{orange!20}0.040 & \cellcolor{orange!20}0.046 & \cellcolor{orange!20}0.041 \\
\hline
\multicolumn{4}{l}{\textit{Test Ratio: 80\%}} \\
Global Mean     & 0.140 & 0.115 & 0.093 & 0.115 & 0.131 & 0.132 & 0.139 \\
Mean of Means   & 0.119 & 0.097 & 0.094 & 0.099 & 0.109 & 0.114 & 0.123 \\
Ours (Profiles) & \cellcolor{orange!20}0.089 & 0.099 & \cellcolor{orange!20}0.090 & 0.096 & 0.082 & 0.111 & 0.117 \\
Ours (Models)   & 0.094 & \cellcolor{orange!20}0.081 & 0.092 & \cellcolor{orange!20}0.088 & \cellcolor{orange!20}0.075 & \cellcolor{orange!20}0.092 & \cellcolor{orange!20}0.095 \\
\hline
\end{tabular}}
\label{tab:modeling_settings}
\end{table}

\begin{table}
\scriptsize
\centering
\caption{\textbf{Performance of Combining Coreset Methods with Our Approach.} As seen, coreset methods and our approach are complementary. Their combination achieves better performance than either method individually.}
\setlength{\tabcolsep}{1.5mm}{
\begin{tabular}{lcccccc}
\toprule
\multirow{2}[4]{*}{Method} & \multicolumn{2}{c}{Overall} & \multicolumn{2}{c}{Acc} & \multicolumn{2}{c}{BART} \\
\cmidrule{2-7}
& RMSE$\downarrow$ & MAE$\downarrow$ & RMSE & MAE & RMSE & MAE \\
\midrule
Ours & 0.193 & 0.090 & 0.074 & 0.052 & 0.459 & 0.299 \\
Random Selection & 0.224 & 0.141 & 0.152 & 0.107 & 0.444 & 0.326 \\
\quad + Ours (Avg) & 0.149 & 0.088 & 0.085 & 0.061 & 0.322 & 0.237 \\
\quad + Ours (Unc) & 0.139 & \cellcolor{orange!20}0.076 & \cellcolor{orange!20}0.069 & \cellcolor{orange!20}0.049 & 0.313 & 0.224 \\
Herding & 0.216 & 0.140 & 0.155 & 0.112 & 0.410 & 0.297 \\
\quad + Ours (Avg) & 0.144 & 0.088 & 0.087 & 0.064 & \cellcolor{orange!20}0.305 & \cellcolor{orange!20}0.220 \\
\quad + Ours (Unc) & 0.140 & \cellcolor{orange!20}0.076 & 0.070 & \cellcolor{orange!20}0.049 & 0.315 & \cellcolor{orange!20}0.220 \\
K-Center Greedy & 0.223 & 0.142 & 0.154 & 0.109 & 0.437 & 0.322 \\
\quad + Ours (Avg) & 0.144 & 0.088 & 0.086 & 0.063 & 0.306 & 0.226 \\
\quad + Ours (Unc) & \cellcolor{orange!20}0.138 & 0.077 & 0.070 & \cellcolor{orange!20}0.049 & 0.313 & 0.224 \\
\bottomrule
\end{tabular}}
\label{tab:coreset_plus_ours}
\end{table}

\subsection{Combine Coreset Method with Our Approach}\label{sec:coreset_plus_ours}

Previous studies~\citep{tinybenchmarks,lmms_eval} select a small set of representative samples as a coreset in a benchmark and evaluate models on the coreset. Our work utilizes known model performance from different benchmarks or models and is complementary to existing methods.

In experiments, we explore two combination approaches. (1) Avg: simply average predictions of coreset methods and PMF; (2) Unc: use uncertainties from MCMC to combine the coreset and PMF predictions. In short, when PMF is confident, we mainly rely on using known performance for prediction, and otherwise, rely on the coreset methods.

Following \citet{lmms_eval}, we use CLIP to generate embeddings for images and BGE-M3 for text, and concatenates them to get the samples embeddings. Based on sample embeddings, we use random, Herding, and K-Center Greedy to select 10\% core samples from benchmarks. 

\tabref{tab:coreset_plus_ours} demonstrates the effectiveness of our approach. More experimental results can be found in the supplementary materials (\secref{sec:app_coreset_plus_ours}).

\subsection{What Can We Tell Based on Vision Encoders?}
Constrained PMF can capture the impact of model and dataset profiles. Here, we present a showcase analysis focusing on the impact of vision encoder type. Specifically, we calculate the dot product between the feature vector of the vision encoder type, $\bsH_m$, and the feature vector of the dataset, $\bsV'_n$. The calculation result measures the influence of a vision encoder on a task. In experiments, CLIP and DINO show improvements on a few datasets, while FNet, SigLIP, and ViT are less effective in comparison. The details are shown in the supplementary material (\secref{sec:app_insight_vision_encoders}).

\subsection{Which Models or Benchmarks are Most Informative to Performance Prediction?}
We assess how representative a model is and how informative a benchmark is, by measuring the RMSE improvements of PMF when we add the full results of a model or dataset. The most informative models and tasks are shown in \figref{fig:insight_informative}. As observed, strong models like GPT-4, Gemini, and InternLM are more representative than weaker models. This is likely because their performance tends to deviate from the average and, being more general, they reliably reflect the difficulty level of various datasets. Interestingly, the text-to-image generation task is particularly informative. In this task, models must select the correct generated image from four candidates, and we observe that strong models, such as GPT-4, perform significantly better than others. This performance gap leads to larger errors in PMF, so including this dataset can significantly improve the PMF model.

\section{Conclusion and Future Work}
Our framework estimates unknown LVLM performances across tasks using PMF, prioritizes evaluations based on uncertainty, and introduces some enhancements to address the sparse data issue. Our study could result in significant savings in development time and computation costs. We highlight two limitations. First, modeling different LVLM evaluation settings, such as 5-shot and chain-of-thought could extend our framework. While some preliminary study is shown in \secref{sec:modeling_settings}, more exploration is left as future work. Second, it is challenging to apply our framework to new models or datasets without any performance data. We discuss this in the supplementary material (\secref{sec:new_models_datasets}).

\section*{Acknowledgments}
We would like to extend our deepest appreciation to Qiyuan Zhang, Caixia Zhou, Weijian Deng, Xingjian Leng, Evan Markou, Yunzhong Hou, Yicong Hong, Taojun Lin, Jiahao Zhang, Sam Bahrami, Shu Zou, Zeyu Zhang, Yang Yang, Jiahao Ma, and all our other lab colleagues for their invaluable support throughout this project. Their collaborative efforts, insightful discussions, and constructive feedback have been crucial in shaping and improving our paper. 

This work was supported by an Australian Research Council (ARC) Linkage grant (project number LP210200931). Qinyu Zhao was supported by OpenAI Researcher Access Program (0000005453) and Google Cloud Research Credits Program (353970026).

\section*{Impact Statement}
This paper presents work whose goal is to advance the field of Machine Learning. There are many potential societal consequences of our work, none which we feel must be specifically highlighted here.

\bibliography{reference}

\begin{thebibliography}{82}
\providecommand{\natexlab}[1]{#1}
\providecommand{\url}[1]{\texttt{#1}}
\expandafter\ifx\csname urlstyle\endcsname\relax
  \providecommand{\doi}[1]{doi: #1}\else
  \providecommand{\doi}{doi: \begingroup \urlstyle{rm}\Url}\fi

\bibitem[Achiam et~al.(2023)Achiam, Adler, Agarwal, Ahmad, Akkaya, Aleman, Almeida, Altenschmidt, Altman, Anadkat, et~al.]{gpt4}
Achiam, J., Adler, S., Agarwal, S., Ahmad, L., Akkaya, I., Aleman, F.~L., Almeida, D., Altenschmidt, J., Altman, S., Anadkat, S., et~al.
\newblock {GPT-4} technical report.
\newblock \emph{arXiv preprint arXiv:2303.08774}, 2023.

\bibitem[Agrawal et~al.(2019)Agrawal, Desai, Wang, Chen, Jain, Johnson, Batra, Parikh, Lee, and Anderson]{nocaps}
Agrawal, H., Desai, K., Wang, Y., Chen, X., Jain, R., Johnson, M., Batra, D., Parikh, D., Lee, S., and Anderson, P.
\newblock {NoCaps}: Novel object captioning at scale.
\newblock In \emph{Proceedings of the IEEE International Conference on Computer Vision}, pp.\  8948--8957, 2019.

\bibitem[Antol et~al.(2015)Antol, Agrawal, Lu, Mitchell, Batra, Zitnick, and Parikh]{antol2015vqa}
Antol, S., Agrawal, A., Lu, J., Mitchell, M., Batra, D., Zitnick, C.~L., and Parikh, D.
\newblock {VQA}: Visual question answering.
\newblock In \emph{Proceedings of the IEEE International Conference on Computer Vision}, pp.\  2425--2433, 2015.

\bibitem[Arevalo et~al.(2017)Arevalo, Solorio, Montes-y G{\'o}mez, and Gonz{\'a}lez]{MMIMDB}
Arevalo, J., Solorio, T., Montes-y G{\'o}mez, M., and Gonz{\'a}lez, F.~A.
\newblock Gated multimodal units for information fusion.
\newblock \emph{arXiv preprint arXiv:1702.01992}, 2017.

\bibitem[Arora et~al.(2019)Arora, Cohen, Hu, and Luo]{dmf}
Arora, S., Cohen, N., Hu, W., and Luo, Y.
\newblock Implicit regularization in deep matrix factorization.
\newblock \emph{Advances in Neural Information Processing Systems}, 32, 2019.

\bibitem[Bai et~al.(2025)Bai, Chen, Liu, Wang, Ge, Song, Dang, Wang, Wang, Tang, et~al.]{bai2025qwen2_5}
Bai, S., Chen, K., Liu, X., Wang, J., Ge, W., Song, S., Dang, K., Wang, P., Wang, S., Tang, J., et~al.
\newblock {Qwen2.5-VL} technical report.
\newblock \emph{arXiv preprint arXiv:2502.13923}, 2025.

\bibitem[Berg et~al.(2017)Berg, Kipf, and Welling]{gcmc}
Berg, R. v.~d., Kipf, T.~N., and Welling, M.
\newblock Graph convolutional matrix completion.
\newblock \emph{arXiv preprint arXiv:1706.02263}, 2017.

\bibitem[Brinkhaus et~al.(2022)Brinkhaus, Zielesny, Steinbeck, and Rajan]{decimer}
Brinkhaus, H.~O., Zielesny, A., Steinbeck, C., and Rajan, K.
\newblock {DECIMER}—hand-drawn molecule images dataset.
\newblock \emph{Journal of Cheminformatics}, 14\penalty0 (1):\penalty0 36, 2022.

\bibitem[Chen et~al.(2023)Chen, Wu, Wang, Su, Chen, Xing, Zhong, Zhang, Zhu, Lu, Li, Luo, Lu, Qiao, and Dai]{internvl}
Chen, Z., Wu, J., Wang, W., Su, W., Chen, G., Xing, S., Zhong, M., Zhang, Q., Zhu, X., Lu, L., Li, B., Luo, P., Lu, T., Qiao, Y., and Dai, J.
\newblock {InternVL}: Scaling up vision foundation models and aligning for generic visual-linguistic tasks.
\newblock \emph{arXiv preprint arXiv:2312.14238}, 2023.

\bibitem[Chen et~al.(2024)Chen, Wang, Tian, Ye, Gao, Cui, Tong, Hu, Luo, Ma, et~al.]{internvl_1_5}
Chen, Z., Wang, W., Tian, H., Ye, S., Gao, Z., Cui, E., Tong, W., Hu, K., Luo, J., Ma, Z., et~al.
\newblock How far are we to {GPT-4V}? closing the gap to commercial multimodal models with open-source suites.
\newblock \emph{arXiv preprint arXiv:2404.16821}, 2024.

\bibitem[Cheng et~al.(2017)Cheng, Han, and Lu]{Resisc45}
Cheng, G., Han, J., and Lu, X.
\newblock Remote sensing image scene classification: Benchmark and state of the art.
\newblock \emph{Proceedings of the IEEE}, 105\penalty0 (10):\penalty0 1865--1883, 2017.

\bibitem[Dai et~al.(2023)Dai, Li, Li, Tiong, Zhao, Wang, Li, Fung, and Hoi]{instructblip}
Dai, W., Li, J., Li, D., Tiong, A. M.~H., Zhao, J., Wang, W., Li, B., Fung, P., and Hoi, S.
\newblock {InstructBLIP}: Towards general-purpose vision-language models with instruction tuning.
\newblock \emph{arXiv preprint arXiv:2305.06500}, 2023.

\bibitem[Deng \& Zheng(2021)Deng and Zheng]{autoeval}
Deng, W. and Zheng, L.
\newblock Are labels always necessary for classifier accuracy evaluation?
\newblock In \emph{Proceedings of the IEEE/CVF Conference on Computer Vision and Pattern Recognition}, pp.\  15069--15078, 2021.

\bibitem[Duan et~al.(2024)Duan, Yang, Qiao, Fang, Chen, Liu, Dong, Zang, Zhang, Wang, et~al.]{VLMEvalKit}
Duan, H., Yang, J., Qiao, Y., Fang, X., Chen, L., Liu, Y., Dong, X., Zang, Y., Zhang, P., Wang, J., et~al.
\newblock {VLMEvalKit}: An open-source toolkit for evaluating large multi-modality models.
\newblock \emph{arXiv preprint arXiv:2407.11691}, 2024.

\bibitem[Fu et~al.(2023)Fu, Chen, Shen, Qin, Zhang, Lin, Yang, Zheng, Li, Sun, et~al.]{fu2023mme}
Fu, C., Chen, P., Shen, Y., Qin, Y., Zhang, M., Lin, X., Yang, J., Zheng, X., Li, K., Sun, X., et~al.
\newblock {MME}: A comprehensive evaluation benchmark for multimodal large language models.
\newblock \emph{arXiv preprint arXiv:2306.13394}, 2023.

\bibitem[Fu et~al.(2024)Fu, Dai, Luo, Li, Ren, Zhang, Wang, Zhou, Shen, Zhang, et~al.]{videomme}
Fu, C., Dai, Y., Luo, Y., Li, L., Ren, S., Zhang, R., Wang, Z., Zhou, C., Shen, Y., Zhang, M., et~al.
\newblock {Video-MME}: The first-ever comprehensive evaluation benchmark of multi-modal llms in video analysis.
\newblock \emph{arXiv preprint arXiv:2405.21075}, 2024.

\bibitem[Goodfellow et~al.(2013)Goodfellow, Erhan, Carrier, Courville, Mirza, Hamner, Cukierski, Tang, Thaler, Lee, et~al.]{goodfellow2013challenges}
Goodfellow, I.~J., Erhan, D., Carrier, P.~L., Courville, A., Mirza, M., Hamner, B., Cukierski, W., Tang, Y., Thaler, D., Lee, D.-H., et~al.
\newblock Challenges in representation learning: A report on three machine learning contests.
\newblock In \emph{Neural information processing: 20th international conference, ICONIP 2013, daegu, korea, november 3-7, 2013. Proceedings, Part III 20}, pp.\  117--124. Springer, 2013.

\bibitem[Guillory et~al.(2021)Guillory, Shankar, Ebrahimi, Darrell, and Schmidt]{doc}
Guillory, D., Shankar, V., Ebrahimi, S., Darrell, T., and Schmidt, L.
\newblock Predicting with confidence on unseen distributions.
\newblock In \emph{Proceedings of the IEEE/CVF Conference on Computer Vision and Pattern Recognition}, pp.\  1134--1144, 2021.

\bibitem[Hao et~al.(2025)Hao, Gu, Wang, Li, Yang, Wang, and Cheng]{emma}
Hao, Y., Gu, J., Wang, H.~W., Li, L., Yang, Z., Wang, L., and Cheng, Y.
\newblock Can {MLLMs} reason in multimodality? {EMMA}: An enhanced multimodal reasoning benchmark.
\newblock \emph{arXiv preprint arXiv:2501.05444}, 2025.

\bibitem[He et~al.(2020)He, Zhang, Mou, Xing, and Xie]{pathvqa}
He, X., Zhang, Y., Mou, L., Xing, E., and Xie, P.
\newblock {PathVQA}: 30000+ questions for medical visual question answering.
\newblock \emph{arXiv preprint arXiv:2003.10286}, 2020.

\bibitem[Hessel et~al.(2022)Hessel, Marasovi{\'c}, Hwang, Lee, Da, Zellers, Mankoff, and Choi]{nycap}
Hessel, J., Marasovi{\'c}, A., Hwang, J.~D., Lee, L., Da, J., Zellers, R., Mankoff, R., and Choi, Y.
\newblock Do androids laugh at electric sheep? humor "understanding" benchmarks from the new yorker caption contest.
\newblock \emph{arXiv preprint arXiv:2209.06293}, 2022.

\bibitem[Hoffman et~al.(2014)Hoffman, Gelman, et~al.]{nuts}
Hoffman, M.~D., Gelman, A., et~al.
\newblock The {No-U-Turn} sampler: Adaptively setting path lengths in {Hamiltonian Monte Carlo}.
\newblock \emph{Journal of Machine Learning Research}, 15\penalty0 (1):\penalty0 1593--1623, 2014.

\bibitem[Huang et~al.(2023)Huang, Bianchi, Yuksekgonul, Montine, and Zou]{openpath}
Huang, Z., Bianchi, F., Yuksekgonul, M., Montine, T., and Zou, J.
\newblock Leveraging medical twitter to build a visual--language foundation model for pathology {AI}.
\newblock \emph{bioRxiv}, pp.\  2023--03, 2023.

\bibitem[Hudson \& Manning(2019)Hudson and Manning]{gqa}
Hudson, D.~A. and Manning, C.~D.
\newblock {GQA}: A new dataset for real-world visual reasoning and compositional question answering.
\newblock In \emph{Proceedings of the IEEE/CVF conference on computer vision and pattern recognition}, pp.\  6700--6709, 2019.

\bibitem[Hwang \& Shwartz(2023)Hwang and Shwartz]{hwang2023memecap}
Hwang, E. and Shwartz, V.
\newblock {MemeCap}: A dataset for captioning and interpreting memes.
\newblock \emph{arXiv preprint arXiv:2305.13703}, 2023.

\bibitem[Karamcheti et~al.(2024)Karamcheti, Nair, Balakrishna, Liang, Kollar, and Sadigh]{prismatic}
Karamcheti, S., Nair, S., Balakrishna, A., Liang, P., Kollar, T., and Sadigh, D.
\newblock {Prismatic VLMs}: Investigating the design space of visually-conditioned language models.
\newblock \emph{arXiv preprint arXiv:2402.07865}, 2024.

\bibitem[Kiela et~al.(2020)Kiela, Firooz, Mohan, Goswami, Singh, Ringshia, and Testuggine]{kiela2020hateful}
Kiela, D., Firooz, H., Mohan, A., Goswami, V., Singh, A., Ringshia, P., and Testuggine, D.
\newblock The hateful memes challenge: Detecting hate speech in multi-modal memes.
\newblock \emph{Advances in Neural Information Processing Systems}, 33:\penalty0 2611--2624, 2020.

\bibitem[Krishna et~al.(2017)Krishna, Zhu, Groth, Johnson, Hata, Kravitz, Chen, Kalantidis, Li, Shamma, et~al.]{vg}
Krishna, R., Zhu, Y., Groth, O., Johnson, J., Hata, K., Kravitz, J., Chen, S., Kalantidis, Y., Li, L.-J., Shamma, D.~A., et~al.
\newblock Visual genome: Connecting language and vision using crowdsourced dense image annotations.
\newblock \emph{International Journal of Computer Vision}, 123:\penalty0 32--73, 2017.

\bibitem[Lau et~al.(2018)Lau, Gayen, Ben~Abacha, and Demner-Fushman]{vqarad}
Lau, J.~J., Gayen, S., Ben~Abacha, A., and Demner-Fushman, D.
\newblock A dataset of clinically generated visual questions and answers about radiology images.
\newblock \emph{Scientific Data}, 5\penalty0 (1):\penalty0 1--10, 2018.

\bibitem[Leiva et~al.(2020)Leiva, Hota, and Oulasvirta]{enrico}
Leiva, L.~A., Hota, A., and Oulasvirta, A.
\newblock {ENRICO}: A high-quality dataset for topic modeling of mobile ui designs.
\newblock \emph{Proc. MobileHCI extended abstracts}, 2020.

\bibitem[Lewandowski et~al.(2009)Lewandowski, Kurowicka, and Joe]{lkj}
Lewandowski, D., Kurowicka, D., and Joe, H.
\newblock Generating random correlation matrices based on vines and extended onion method.
\newblock \emph{Journal of Multivariate Analysis}, 100\penalty0 (9):\penalty0 1989--2001, 2009.

\bibitem[Li et~al.(2023{\natexlab{a}})Li, Ge, Ge, Wang, Wang, Zhang, and Shan]{seed2}
Li, B., Ge, Y., Ge, Y., Wang, G., Wang, R., Zhang, R., and Shan, Y.
\newblock {SEED-Bench-2}: Benchmarking multimodal large language models.
\newblock \emph{arXiv preprint arXiv:2311.17092}, 2023{\natexlab{a}}.

\bibitem[Li \& Lu(2024)Li and Lu]{li2024survey}
Li, J. and Lu, W.
\newblock A survey on benchmarks of multimodal large language models.
\newblock \emph{arXiv preprint arXiv:2408.08632}, 2024.

\bibitem[Li et~al.(2023{\natexlab{b}})Li, Du, Zhou, Wang, Zhao, and Wen]{pope_benchmark}
Li, Y., Du, Y., Zhou, K., Wang, J., Zhao, W.~X., and Wen, J.-R.
\newblock Evaluating object hallucination in large vision-language models.
\newblock \emph{arXiv preprint arXiv:2305.10355}, 2023{\natexlab{b}}.

\bibitem[Liang et~al.(2024)Liang, Goindani, Chafekar, Mathur, Yu, Salakhutdinov, and Morency]{hemm}
Liang, P.~P., Goindani, A., Chafekar, T., Mathur, L., Yu, H., Salakhutdinov, R., and Morency, L.-P.
\newblock {HEMM}: Holistic evaluation of multimodal foundation models.
\newblock \emph{arXiv preprint arXiv:2407.03418}, 2024.

\bibitem[Liu et~al.(2021)Liu, Zhan, Xu, Ma, Yang, and Wu]{slake}
Liu, B., Zhan, L.-M., Xu, L., Ma, L., Yang, Y., and Wu, X.-M.
\newblock {SLAKE}: A semantically-labeled knowledge-enhanced dataset for medical visual question answering.
\newblock In \emph{2021 IEEE 18th International Symposium on Biomedical Imaging (ISBI)}, pp.\  1650--1654. IEEE, 2021.

\bibitem[Liu et~al.(2023{\natexlab{a}})Liu, Li, Li, and Lee]{llava_1_5}
Liu, H., Li, C., Li, Y., and Lee, Y.~J.
\newblock Improved baselines with visual instruction tuning.
\newblock \emph{arXiv preprint arXiv:2310.03744}, 2023{\natexlab{a}}.

\bibitem[Liu et~al.(2024)Liu, Li, Wu, and Lee]{llava_v1}
Liu, H., Li, C., Wu, Q., and Lee, Y.~J.
\newblock Visual instruction tuning.
\newblock \emph{Advances in Neural Information Processing Systems}, 36, 2024.

\bibitem[Liu et~al.(2023{\natexlab{b}})Liu, Duan, Zhang, Li, Zhang, Zhao, Yuan, Wang, He, Liu, et~al.]{mmbench}
Liu, Y., Duan, H., Zhang, Y., Li, B., Zhang, S., Zhao, W., Yuan, Y., Wang, J., He, C., Liu, Z., et~al.
\newblock {MMBench}: Is your multi-modal model an all-around player?
\newblock \emph{arXiv preprint arXiv:2307.06281}, 2023{\natexlab{b}}.

\bibitem[Lu et~al.(2024)Lu, Liu, Zhang, Wang, Dong, Liu, Sun, Ren, Li, Yang, et~al.]{lu2024deepseek}
Lu, H., Liu, W., Zhang, B., Wang, B., Dong, K., Liu, B., Sun, J., Ren, T., Li, Z., Yang, H., et~al.
\newblock {DeepSeek-VL}: towards real-world vision-language understanding.
\newblock \emph{arXiv preprint arXiv:2403.05525}, 2024.

\bibitem[Lu et~al.(2022)Lu, Mishra, Xia, Qiu, Chang, Zhu, Tafjord, Clark, and Kalyan]{scienceqa}
Lu, P., Mishra, S., Xia, T., Qiu, L., Chang, K.-W., Zhu, S.-C., Tafjord, O., Clark, P., and Kalyan, A.
\newblock Learn to explain: Multimodal reasoning via thought chains for science question answering.
\newblock \emph{Advances in Neural Information Processing Systems}, 35:\penalty0 2507--2521, 2022.

\bibitem[Marino et~al.(2019)Marino, Rastegari, Farhadi, and Mottaghi]{okvqa}
Marino, K., Rastegari, M., Farhadi, A., and Mottaghi, R.
\newblock {OK-VQA}: A visual question answering benchmark requiring external knowledge.
\newblock In \emph{Proceedings of the IEEE/cvf conference on computer vision and pattern recognition}, pp.\  3195--3204, 2019.

\bibitem[Mnih \& Salakhutdinov(2007)Mnih and Salakhutdinov]{pmf}
Mnih, A. and Salakhutdinov, R.~R.
\newblock Probabilistic matrix factorization.
\newblock \emph{Advances in Neural Information Processing Systems}, 20, 2007.

\bibitem[Neal(2011)]{hmc}
Neal, R.
\newblock Handbook of {Markov chain Monte Carlo}, volume 2, chapter {MCMC} using hamiltonian dynamics, 2011.

\bibitem[OpenAI(2024)]{O1}
OpenAI.
\newblock {GPT-4o}.
\newblock \url{https://openai.com/o1/}, 2024.
\newblock Accessed: 2025-05-29.

\bibitem[Perlitz et~al.(2023)Perlitz, Bandel, Gera, Arviv, Ein-Dor, Shnarch, Slonim, Shmueli-Scheuer, and Choshen]{efficient_benchmarking}
Perlitz, Y., Bandel, E., Gera, A., Arviv, O., Ein-Dor, L., Shnarch, E., Slonim, N., Shmueli-Scheuer, M., and Choshen, L.
\newblock Efficient benchmarking (of language models).
\newblock \emph{arXiv preprint arXiv:2308.11696}, 2023.

\bibitem[Polo et~al.(2024)Polo, Weber, Choshen, Sun, Xu, and Yurochkin]{tinybenchmarks}
Polo, F.~M., Weber, L., Choshen, L., Sun, Y., Xu, G., and Yurochkin, M.
\newblock {tinyBenchmarks}: Evaluating {LLMs} with fewer examples.
\newblock \emph{arXiv preprint arXiv:2402.14992}, 2024.

\bibitem[Prabhu et~al.(2024)Prabhu, Udandarao, Torr, Bethge, Bibi, and Albanie]{lifelong}
Prabhu, A., Udandarao, V., Torr, P., Bethge, M., Bibi, A., and Albanie, S.
\newblock Lifelong benchmarks: Efficient model evaluation in an era of rapid progress.
\newblock \emph{arXiv preprint arXiv:2402.19472}, 2024.

\bibitem[Reid et~al.(2024)Reid, Savinov, Teplyashin, Lepikhin, Lillicrap, Alayrac, Soricut, Lazaridou, Firat, Schrittwieser, et~al.]{gemini15}
Reid, M., Savinov, N., Teplyashin, D., Lepikhin, D., Lillicrap, T., Alayrac, J.-b., Soricut, R., Lazaridou, A., Firat, O., Schrittwieser, J., et~al.
\newblock Gemini 1.5: Unlocking multimodal understanding across millions of tokens of context.
\newblock \emph{arXiv preprint arXiv:2403.05530}, 2024.

\bibitem[Rodriguez et~al.(2021)Rodriguez, Barrow, Hoyle, Lalor, Jia, and Boyd-Graber]{irt_eval}
Rodriguez, P., Barrow, J., Hoyle, A.~M., Lalor, J.~P., Jia, R., and Boyd-Graber, J.
\newblock Evaluation examples are not equally informative: How should that change {NLP} leaderboards?
\newblock In \emph{Proceedings of the 59th Annual Meeting of the Association for Computational Linguistics and the 11th International Joint Conference on Natural Language Processing (Volume 1: Long Papers)}, pp.\  4486--4503, 2021.

\bibitem[Salakhutdinov \& Mnih(2008)Salakhutdinov and Mnih]{bpmf}
Salakhutdinov, R. and Mnih, A.
\newblock Bayesian probabilistic matrix factorization using {Markov chain Monte Carlo}.
\newblock In \emph{Proceedings of the 25th International Conference on Machine Learning}, pp.\  880--887, 2008.

\bibitem[Schaeffer et~al.(2024)Schaeffer, Valentine, Bailey, Chua, Eyzaguirre, Durante, Benton, Miranda, Sleight, Hughes, et~al.]{universal}
Schaeffer, R., Valentine, D., Bailey, L., Chua, J., Eyzaguirre, C., Durante, Z., Benton, J., Miranda, B., Sleight, H., Hughes, J., et~al.
\newblock When do universal image jailbreaks transfer between vision-language models?
\newblock \emph{arXiv preprint arXiv:2407.15211}, 2024.

\bibitem[Sharma et~al.(2020)Sharma, Paka, Bhageria, Das, Poria, Chakraborty, and Gamb{\"a}ck]{chhavi2020memotion}
Sharma, C., Paka, Scott, W., Bhageria, D., Das, A., Poria, S., Chakraborty, T., and Gamb{\"a}ck, B.
\newblock {Task Report: Memotion Analysis 1.0 @SemEval 2020: The Visuo-Lingual Metaphor!}
\newblock In \emph{Proceedings of the 14th International Workshop on Semantic Evaluation ({S}em{E}val-2020)}, Barcelona, Spain, Sep 2020. Association for Computational Linguistics.

\bibitem[Song et~al.(2020)Song, Tan, Qin, Lu, and Liu]{mpnet}
Song, K., Tan, X., Qin, T., Lu, J., and Liu, T.-Y.
\newblock {MPNet}: Masked and permuted pre-training for language understanding.
\newblock \emph{Advances in Neural Information Processing Systems}, 33:\penalty0 16857--16867, 2020.

\bibitem[Suhr et~al.(2017)Suhr, Lewis, Yeh, and Artzi]{nlvr}
Suhr, A., Lewis, M., Yeh, J., and Artzi, Y.
\newblock A corpus of natural language for visual reasoning.
\newblock In \emph{Proceedings of the 55th Annual Meeting of the Association for Computational Linguistics (Volume 2: Short Papers)}, pp.\  217--223, 2017.

\bibitem[Suhr et~al.(2018)Suhr, Zhou, Zhang, Zhang, Bai, and Artzi]{nlvr2}
Suhr, A., Zhou, S., Zhang, A., Zhang, I., Bai, H., and Artzi, Y.
\newblock A corpus for reasoning about natural language grounded in photographs.
\newblock \emph{arXiv preprint arXiv:1811.00491}, 2018.

\bibitem[Team et~al.(2023)Team, Anil, Borgeaud, Wu, Alayrac, Yu, Soricut, Schalkwyk, Dai, Hauth, et~al.]{team2023gemini}
Team, G., Anil, R., Borgeaud, S., Wu, Y., Alayrac, J.-B., Yu, J., Soricut, R., Schalkwyk, J., Dai, A.~M., Hauth, A., et~al.
\newblock {Gemini}: A family of highly capable multimodal models.
\newblock \emph{arXiv preprint arXiv:2312.11805}, 2023.

\bibitem[Thrush et~al.(2022)Thrush, Jiang, Bartolo, Singh, Williams, Kiela, and Ross]{winoground}
Thrush, T., Jiang, R., Bartolo, M., Singh, A., Williams, A., Kiela, D., and Ross, C.
\newblock Winoground: Probing vision and language models for visio-linguistic compositionality.
\newblock In \emph{Proceedings of the IEEE/CVF Conference on Computer Vision and Pattern Recognition}, pp.\  5238--5248, 2022.

\bibitem[Tong et~al.(2024{\natexlab{a}})Tong, Brown, Wu, Woo, Middepogu, Akula, Yang, Yang, Iyer, Pan, et~al.]{cambrian}
Tong, S., Brown, E., Wu, P., Woo, S., Middepogu, M., Akula, S.~C., Yang, J., Yang, S., Iyer, A., Pan, X., et~al.
\newblock Cambrian-1: A fully open, vision-centric exploration of multimodal {LLMs}.
\newblock \emph{arXiv preprint arXiv:2406.16860}, 2024{\natexlab{a}}.

\bibitem[Tong et~al.(2024{\natexlab{b}})Tong, Liu, Zhai, Ma, LeCun, and Xie]{tong2024eyes}
Tong, S., Liu, Z., Zhai, Y., Ma, Y., LeCun, Y., and Xie, S.
\newblock Eyes wide shut? exploring the visual shortcomings of multimodal {LLMs}.
\newblock In \emph{Proceedings of the IEEE/CVF Conference on Computer Vision and Pattern Recognition}, pp.\  9568--9578, 2024{\natexlab{b}}.

\bibitem[Van~Horn et~al.(2018)Van~Horn, Mac~Aodha, Song, Cui, Sun, Shepard, Adam, Perona, and Belongie]{inat}
Van~Horn, G., Mac~Aodha, O., Song, Y., Cui, Y., Sun, C., Shepard, A., Adam, H., Perona, P., and Belongie, S.
\newblock The {iNaturalist} species classification and detection dataset.
\newblock In \emph{Proceedings of the IEEE/CVF Conference on Computer Vision and Pattern Recognition}, pp.\  8769--8778, 2018.

\bibitem[Wang et~al.(2021)Wang, Li, Zhou, Chen, Grossman, and Li]{screen2words}
Wang, B., Li, G., Zhou, X., Chen, Z., Grossman, T., and Li, Y.
\newblock {Screen2Words}: Automatic mobile {UI} summarization with multimodal learning.
\newblock In \emph{The 34th Annual ACM Symposium on User Interface Software and Technology}, pp.\  498--510, 2021.

\bibitem[Wang et~al.(2024{\natexlab{a}})Wang, Pan, Shi, Lu, Ren, Zhou, Zhan, and Li]{mathvision}
Wang, K., Pan, J., Shi, W., Lu, Z., Ren, H., Zhou, A., Zhan, M., and Li, H.
\newblock Measuring multimodal mathematical reasoning with {MathVision} dataset.
\newblock \emph{Advances in Neural Information Processing Systems}, 37:\penalty0 95095--95169, 2024{\natexlab{a}}.

\bibitem[Wang et~al.(2024{\natexlab{b}})Wang, Bai, Tan, Wang, Fan, Bai, Chen, Liu, Wang, Ge, et~al.]{wang2024qwen2}
Wang, P., Bai, S., Tan, S., Wang, S., Fan, Z., Bai, J., Chen, K., Liu, X., Wang, J., Ge, W., et~al.
\newblock {Qwen2-VL}: Enhancing vision-language model's perception of the world at any resolution.
\newblock \emph{arXiv preprint arXiv:2409.12191}, 2024{\natexlab{b}}.

\bibitem[Wu et~al.(2024)Wu, Li, Chen, and Li]{longvideobench}
Wu, H., Li, D., Chen, B., and Li, J.
\newblock {LongVideoBench}: A benchmark for long-context interleaved video-language understanding.
\newblock \emph{Advances in Neural Information Processing Systems}, 37:\penalty0 28828--28857, 2024.

\bibitem[Yang \& Newsam(2010)Yang and Newsam]{UCMerced}
Yang, Y. and Newsam, S.
\newblock Bag-of-visual-words and spatial extensions for land-use classification.
\newblock In \emph{Proceedings of the 18th SIGSPATIAL international conference on advances in geographic information systems}, pp.\  270--279, 2010.

\bibitem[Yang et~al.(2024)Yang, Wang, Chen, Dai, and Zheng]{bbs}
Yang, Y., Wang, W., Chen, Z., Dai, J., and Zheng, L.
\newblock Bounding box stability against feature dropout reflects detector generalization across environments.
\newblock \emph{arXiv preprint arXiv:2403.13803}, 2024.

\bibitem[Ye et~al.(2023)Ye, Xu, Ye, Yan, Liu, Qian, Zhang, Huang, and Zhou]{ye2023mplug2}
Ye, Q., Xu, H., Ye, J., Yan, M., Liu, H., Qian, Q., Zhang, J., Huang, F., and Zhou, J.
\newblock {mPLUG-Owl2}: Revolutionizing multi-modal large language model with modality collaboration.
\newblock \emph{arXiv preprint arXiv:2311.04257}, 2023.

\bibitem[Yosef et~al.(2023)Yosef, Bitton, and Shahaf]{irfl}
Yosef, R., Bitton, Y., and Shahaf, D.
\newblock {IRFL}: Image recognition of figurative language.
\newblock \emph{arXiv preprint arXiv:2303.15445}, 2023.

\bibitem[Young et~al.(2014)Young, Lai, Hodosh, and Hockenmaier]{flickr30k}
Young, P., Lai, A., Hodosh, M., and Hockenmaier, J.
\newblock From image descriptions to visual denotations: New similarity metrics for semantic inference over event descriptions.
\newblock \emph{Transactions of the Association for Computational Linguistics}, 2:\penalty0 67--78, 2014.

\bibitem[Yuan et~al.(2021)Yuan, Neubig, and Liu]{bartscore}
Yuan, W., Neubig, G., and Liu, P.
\newblock {BARTScore}: Evaluating generated text as text generation.
\newblock \emph{Advances in Neural Information Processing Systems}, 34:\penalty0 27263--27277, 2021.

\bibitem[Yue et~al.(2024)Yue, Ni, Zhang, Zheng, Liu, Zhang, Stevens, Jiang, Ren, Sun, et~al.]{mmmu}
Yue, X., Ni, Y., Zhang, K., Zheng, T., Liu, R., Zhang, G., Stevens, S., Jiang, D., Ren, W., Sun, Y., et~al.
\newblock {MMMU}: A massive multi-discipline multimodal understanding and reasoning benchmark for expert {AGI}.
\newblock In \emph{Proceedings of the IEEE/CVF Conference on Computer Vision and Pattern Recognition}, pp.\  9556--9567, 2024.

\bibitem[Zellers et~al.(2019)Zellers, Bisk, Farhadi, and Choi]{zellers2019vcr}
Zellers, R., Bisk, Y., Farhadi, A., and Choi, Y.
\newblock From recognition to cognition: Visual commonsense reasoning.
\newblock In \emph{Proceedings of the IEEE/CVF Conference on Computer Vision and Pattern Recognition}, June 2019.

\bibitem[Zhang et~al.(2024{\natexlab{a}})Zhang, Du, Chen, Liang, Luo, Zheng, Zhu, Cheng, Xu, Guo, et~al.]{cmmmu}
Zhang, G., Du, X., Chen, B., Liang, Y., Luo, T., Zheng, T., Zhu, K., Cheng, Y., Xu, C., Guo, S., et~al.
\newblock {CMMMU}: A chinese massive multi-discipline multi-modal understanding benchmark.
\newblock \emph{arXiv preprint arXiv:2401.11944}, 2024{\natexlab{a}}.

\bibitem[Zhang et~al.(2024{\natexlab{b}})Zhang, Li, Zhang, Pu, Cahyono, Hu, Liu, Zhang, Yang, Li, et~al.]{lmms_eval}
Zhang, K., Li, B., Zhang, P., Pu, F., Cahyono, J.~A., Hu, K., Liu, S., Zhang, Y., Yang, J., Li, C., et~al.
\newblock {LMMs-Eval}: Reality check on the evaluation of large multimodal models.
\newblock \emph{arXiv preprint arXiv:2407.12772}, 2024{\natexlab{b}}.

\bibitem[Zhang et~al.(2024{\natexlab{c}})Zhang, Lyu, Liu, and Ma]{zhang2024collaborative}
Zhang, Q., Lyu, F., Liu, X., and Ma, C.
\newblock Collaborative performance prediction for large language models.
\newblock \emph{arXiv preprint arXiv:2407.01300}, 2024{\natexlab{c}}.

\bibitem[Zhang et~al.(2019)Zhang, Kishore, Wu, Weinberger, and Artzi]{bertscore}
Zhang, T., Kishore, V., Wu, F., Weinberger, K.~Q., and Artzi, Y.
\newblock {BERTScore}: Evaluating text generation with {BERT}.
\newblock \emph{arXiv preprint arXiv:1904.09675}, 2019.

\bibitem[Zhang et~al.(2025)Zhang, Zhao, Fang, Li, Liu, Min, Duan, Chen, and Zhai]{zhang2025redundancy}
Zhang, Z., Zhao, X., Fang, X., Li, C., Liu, X., Min, X., Duan, H., Chen, K., and Zhai, G.
\newblock Redundancy principles for {MLLMs} benchmarks.
\newblock \emph{arXiv preprint arXiv:2501.13953}, 2025.

\bibitem[Zheng et~al.(2023)Zheng, Chiang, Sheng, Zhuang, Wu, Zhuang, Lin, Li, Li, Xing, et~al.]{llm_as_judge}
Zheng, L., Chiang, W.-L., Sheng, Y., Zhuang, S., Wu, Z., Zhuang, Y., Lin, Z., Li, Z., Li, D., Xing, E., et~al.
\newblock Judging {LLM-as-a-judge} with {MT-bench} and chatbot arena.
\newblock \emph{Advances in Neural Information Processing Systems}, 36:\penalty0 46595--46623, 2023.

\bibitem[Zhu et~al.(2023)Zhu, Chen, Shen, Li, and Elhoseiny]{minigpt4}
Zhu, D., Chen, J., Shen, X., Li, X., and Elhoseiny, M.
\newblock {MiniGPT-4}: Enhancing vision-language understanding with advanced large language models.
\newblock \emph{arXiv preprint arXiv:2304.10592}, 2023.

\bibitem[Zhu et~al.(2024)Zhu, Zang, Jia, Wu, Fang, Li, Guo, Zheng, Li, Wu, et~al.]{lime}
Zhu, K., Zang, Q., Jia, S., Wu, S., Fang, F., Li, Y., Guo, S., Zheng, T., Li, B., Wu, H., et~al.
\newblock {LIME-M}: Less is more for evaluation of {MLLMs}.
\newblock \emph{arXiv preprint arXiv:2409.06851}, 2024.

\bibitem[Zohar et~al.(2024)Zohar, Wang, Dubois, Mehta, Xiao, Hansen-Estruch, Yu, Wang, Juefei-Xu, Zhang, et~al.]{zohar2024apollo}
Zohar, O., Wang, X., Dubois, Y., Mehta, N., Xiao, T., Hansen-Estruch, P., Yu, L., Wang, X., Juefei-Xu, F., Zhang, N., et~al.
\newblock Apollo: An exploration of video understanding in large multimodal models.
\newblock \emph{arXiv preprint arXiv:2412.10360}, 2024.

\end{thebibliography}
\bibliographystyle{icml2025}

\cleardoublepage

\appendix

\section{Estimation of Evaluation Cost}\label{asec:eval_cost}
To estimate evaluation cost of LVLMs, we evaluate a representative model, Qwen2.5-VL-Instruct 7B~\citep{wang2024qwen2} based on LMMs-Eval~\citep{lmms_eval} with one A100 GPU. The results are shown in \tabref{tab:eval_efficiency}.

As shown in the first and second columns, although vllm significantly accelerate the evaluation, the computational cost is still high. Assuming that we are comparing 5 7B models on 10 similar-scale benchmarks, the evaluation will take around 21.7 hours. Besides, in our experiments, vllm usually leads to slight performance decrease.

LMMs-Lite (a coreset method) significantly reduces evaluation cost, indicating reducing evaluation cost is still valuable in practice. As shown in \secref{sec:coreset_plus_ours}, coreset methods may get inaccurate results, while our method improves them.

\section{Comprehensive Evaluation of LVLMs}\label{asec:evaluating_details}

We provide a comprehensive overview of the datasets and LVLMs used in our study. Detailed dataset information can be found in \tabref{tab:dataset_info} and \ref{tab:dataset_metric}, while the model profiles are presented in Tables \ref{tab:model_info} and \ref{tab:model_info_2}.

\begin{figure*}
\centering
\includegraphics[width=0.98\textwidth,trim=0 0 230 0,clip]{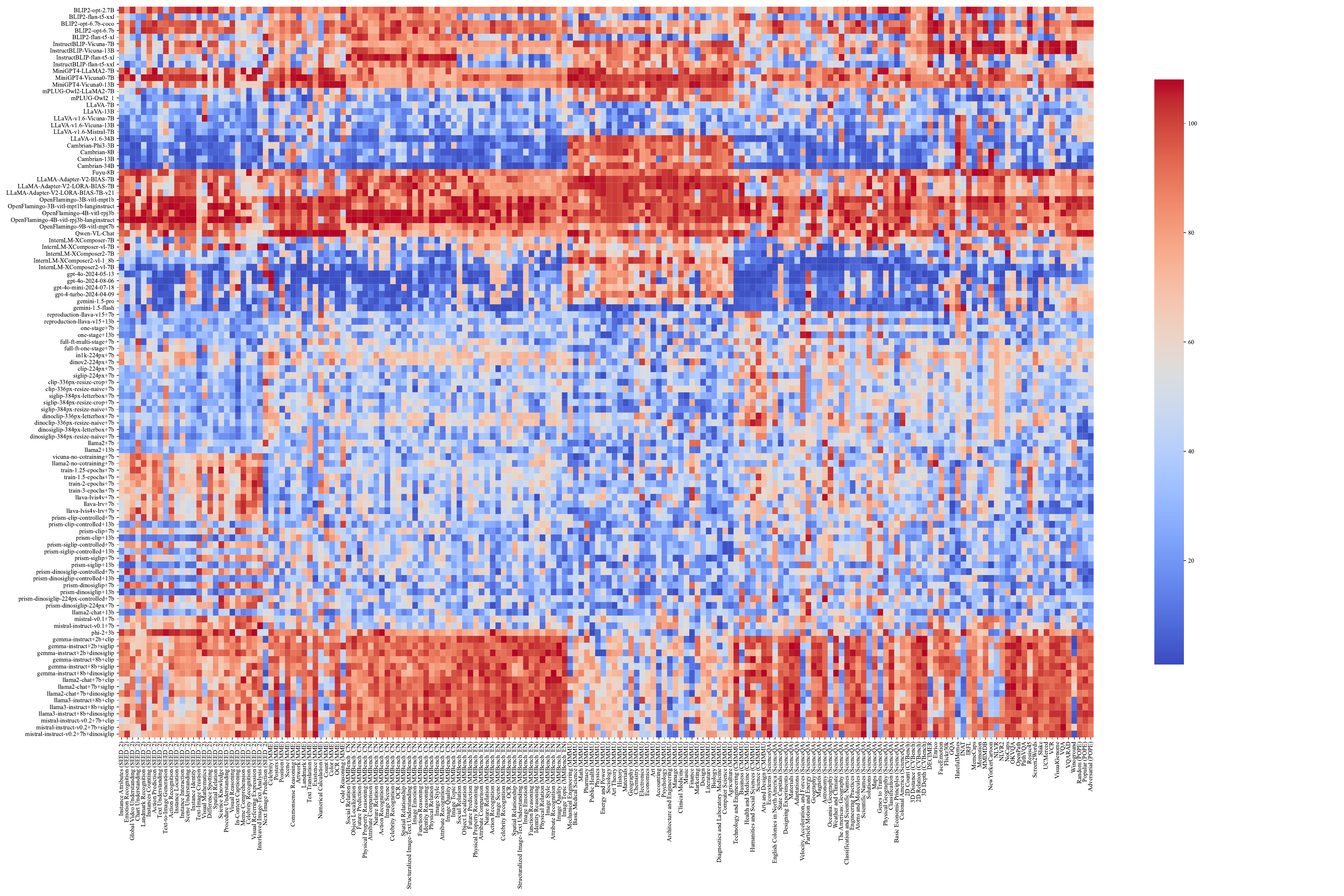}
  \caption{\textbf{Heatmap of Model Rankings on Each Dataset.}} 
  \label{fig:model_perf}
\end{figure*}

\begin{figure*}
\centering
\includegraphics[width=0.98\textwidth,trim=0 0 200 0,clip]{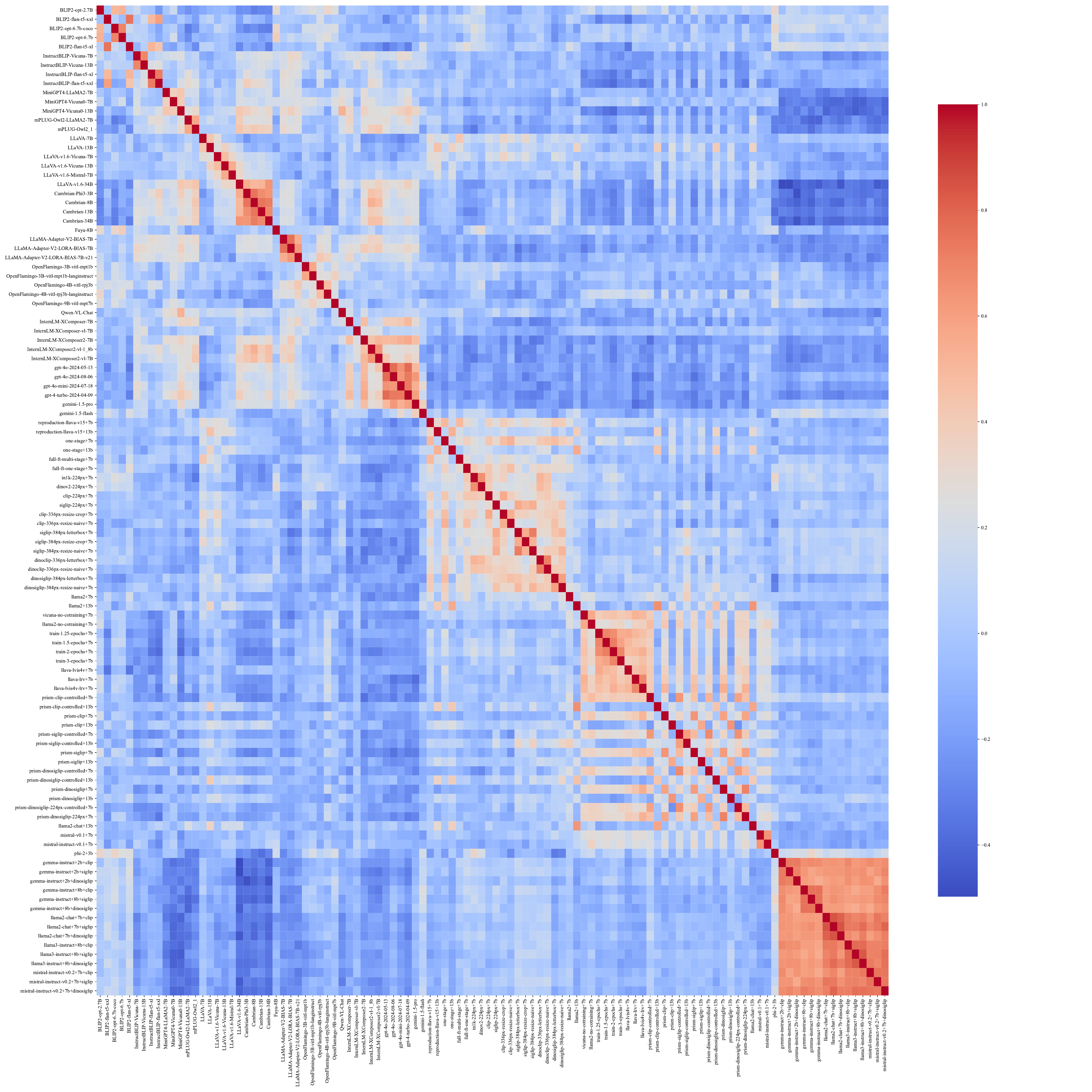}
  \caption{\textbf{Heatmap of Correlation in Model Ranking.}} 
  \label{fig:model_corr}
\end{figure*}

\begin{figure*}
\centering
\includegraphics[width=0.98\textwidth,trim=0 0 200 0,clip]{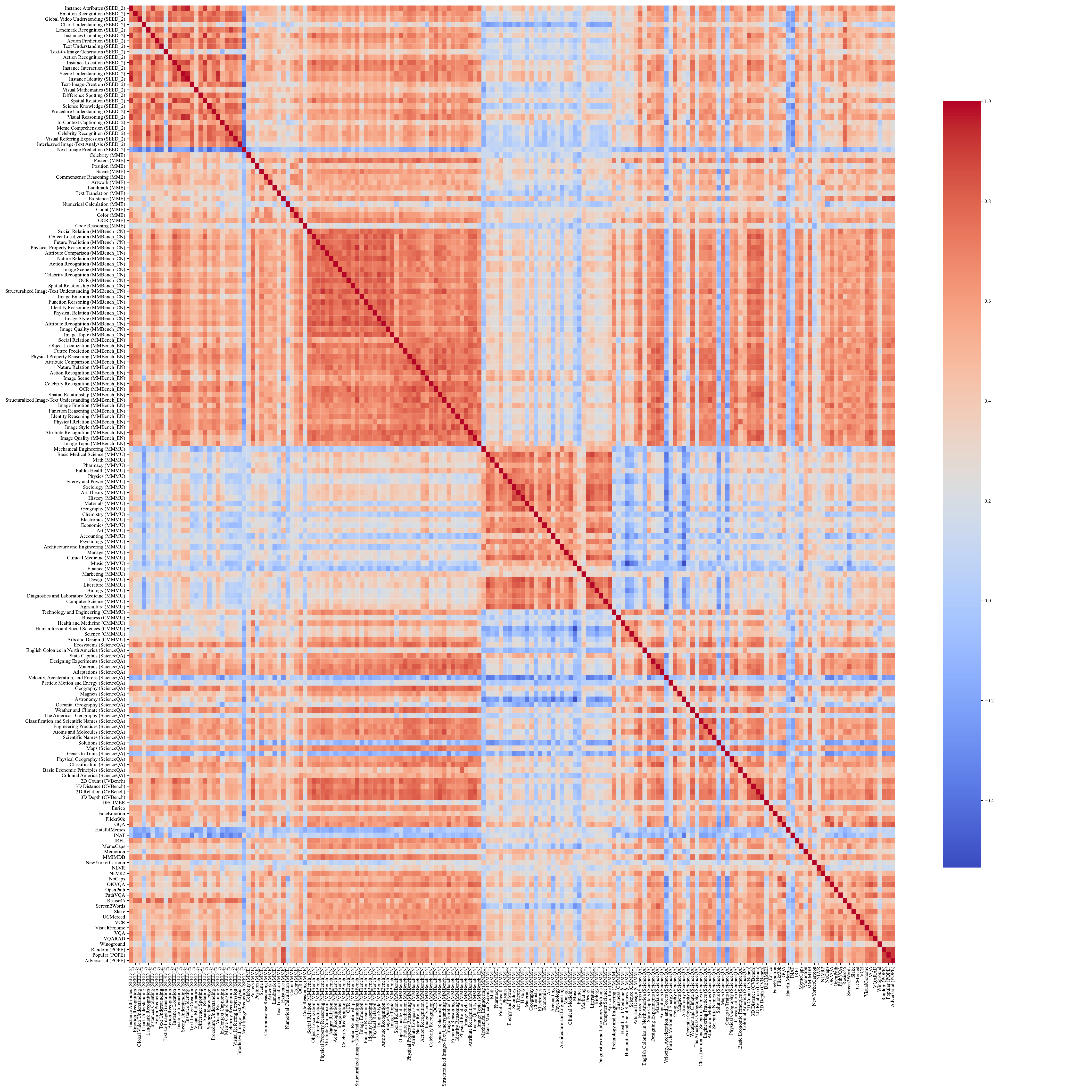}
  \caption{\textbf{Heatmap of Ranking Correlation on Datasets.}} 
  \label{fig:dataset_corr}
\end{figure*}

A heatmap illustrating the model ranking across datasets is shown in \figref{fig:model_perf}. Additionally, the correlation analysis of performance scores is illustrated in \figref{fig:model_corr} and \ref{fig:dataset_corr}. Notably, even within the same model family, such as the LLaVA series, the rankings between models do not exhibit a strong correlation. Datasets tend to have much more consistent ranking correlations, suggesting that models performing well on one dataset are likely to rank highly on others.

\begin{figure*}
\centering
\includegraphics[width=0.98\textwidth,trim=0 20 0 0,clip]{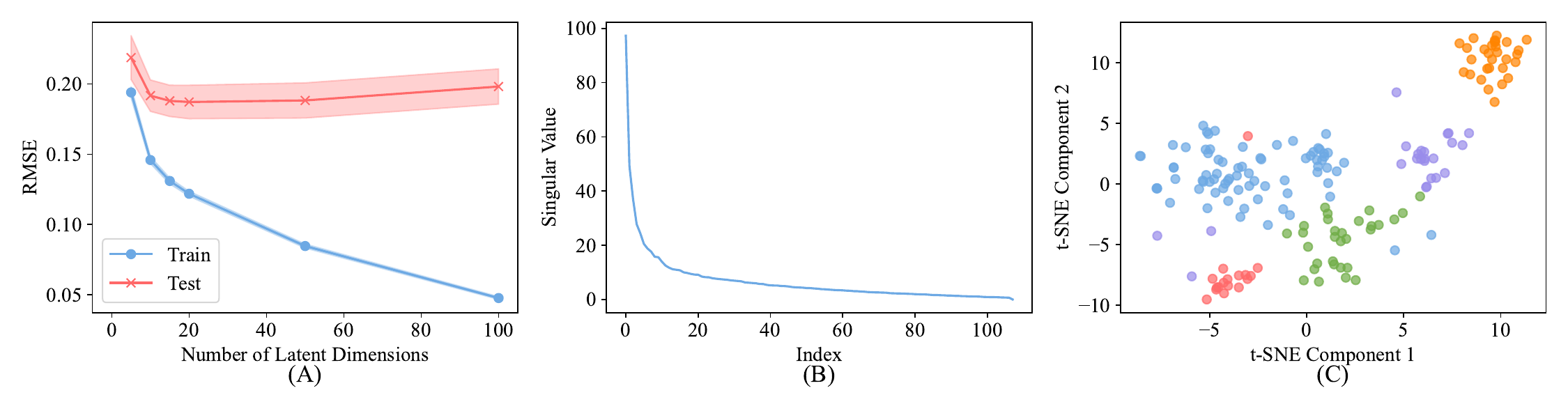}
  \caption{\textbf{Low-Rank Property of the Score Matrix.} (A) RMSE on the test set for PMF stabilizes when the latent dimension exceeds 15. (B) The top singular values of the performance matrix are significantly larger than the others. (C) t-SNE visualization of dataset clusters.} 
  \label{fig:insight_low_rank}
\end{figure*}

Besides, we investigate the impact of different latent dimensions in the PMF models and find that a relatively small latent dimension, around 10, is sufficient. As shown in \figref{fig:insight_low_rank}, increasing the latent dimension reduces the RMSE on the training data to zero due to overfitting, but it does not lead to significant improvements in RMSE on the testing data. Additionally, when we extract the singular values of the score matrix, we observe that the top singular values are much larger than the rest, indicating that most of the information is captured by a few dimensions. This suggests a high degree of similarity in performance scores. 

\section{Further Experimental Results}\label{sec:more_results}
\subsection{Comparing PMF and Deep Learning-Based Methods}\label{sec:app_comp_pmf_dmf}
We evaluate Deep Matrix Factorization (DMF)~\citep{dmf} and Graph Convolutional Matrix Completion (GCMC)~\citep{gcmc} for comparison.

For DMF, we use MSE loss and the Adam optimizer. The learning rate is 1e-3 and the batch size is 256. The embedding dimension of each user or item is 10, which is the same for PMF. We train DMF for 200 epochs and the result of the best epoch is reported.

For GCMC, we refer to the GitHub implementation (https://github.com/riannevdberg/gc-mc). Dropout ratio is 0.7, learning rate is 0.01, hidden units are [500, 75] in 1st and 2nd layers, accumulation method is ``stack'', the number of basis functions is 2, and the model is trained for 200 epochs. We note two main issues when using GCMC:

First, GCMC handles discrete rating levels and treats each rating level as a separate class (see Section 2.3 in \citet{gcmc}), which is not suitable for our setting, because we also use continuous ratings like the BART scores. To address this, we only use LVLM benchmarks with accuracy as the main metric, and discretize accuracy into 101 classes, i.e., {0, ..., 100}.
Second, when training data is sparse, some classes (for example, accuracy = 67) do not occur in training set, leading to running errors in the code.

\tabref{tab:pmf_vs_dmf} summarizes the average results from 10 experiments with different random seeds. As shown, PMF demonstrates superior performance on our dataset compared to DMF and GCMC. We notice that deep matrix completion methods are commonly applied in recommender systems, where there are usually thousands of users and items, with millions of samples. But in our setting, we have 108 models, 176 datasets, and 19K samples. Thus, we build our method on a simple but strong baseline, PMF, rather than neural networks, which are possibly more data-hungry.

\subsection{Detailed Performance Evaluations of PMF and PTF}
We present detailed performance evaluations of PMF in \tabref{tab:detailed_pmf} and PTF in \tabref{tab:detailed_ptf}. As shown, our methods consistently outperform the baselines. In scenarios where performance data is sparse, our enhancements significantly improves the prediction accuracy of PMF.

We also include the following ranking-based metrics.

\textbf{Spearman’s rank correlation.}

\textbf{Kendall rank correlation.}

\textbf{Precision@K.} The proportion of the predicted top 1 model that fall within the top K positions of the ground-truth ranking. For example, our method predicts LLaVA-1.5 is the best on the task and it will be correct if LLaVA-1.5 is within top 5 of the ground-truth ranking for Precision@5.

\textbf{Recall@K.} The proportion of the ground-truth best ones that is correctly retrieved within our top K predicted results. For instance, if LLaVA-1.5 is the best model, it must be within the top 3 predictions of our method for Recall@3.

As shown in \tabref{tab:rank_performance}, ranking-based metrics also supports the advantage of our methods. Baseline methods typically achieve very high precision but low recall, while our methods provide more balanced precision and recall, as well as improved rank correlation. 

``Unc 50\%'' and ``Unc 30\%'' mean keeping the 50\% or 30\% most confident predictions based on our estimated uncertainty, which further improve the estimation accuracy. But note that this may lead to fewer predicted results.

\subsection{Combining Coreset Methods with Our Approach}\label{sec:app_coreset_plus_ours}
We sample different ratios of coreset samples from the benchmarks, and the results are shown in \tabref{tab:coreset_with_ours_diff_ratios}.

\subsection{What Can We Tell Based on Vision Encoders?}\label{sec:app_insight_vision_encoders}
\begin{figure*}
\centering
\includegraphics[width=0.98\textwidth,trim=0 10 0 15,clip]{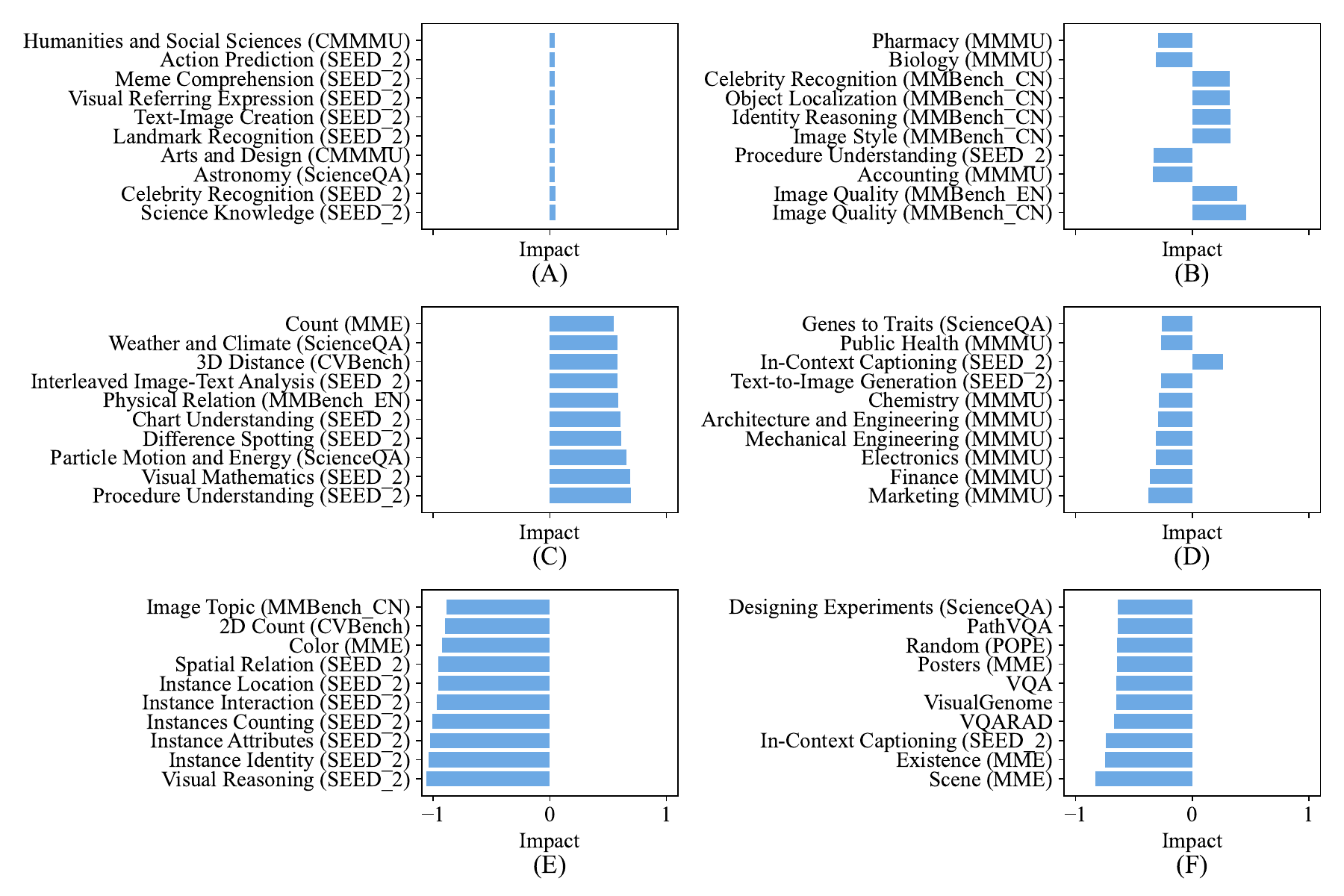}
  \caption{\textbf{Effect Analysis of Vision Encoders on Downstream Tasks.} We evaluate the impact of each vision encoder on downstream tasks by calculating the dot product between the feature vector of the vision encoder and the feature vector of the dataset.}
  \label{fig:insight_profile_info}
\end{figure*}

Detailed results are shown in \figref{fig:insight_profile_info}.

\subsection{Generalization to New Models and Datasets}\label{sec:new_models_datasets}
\begin{figure*}
\centering
\includegraphics[width=0.8\textwidth,trim=0 0 0 0,clip]{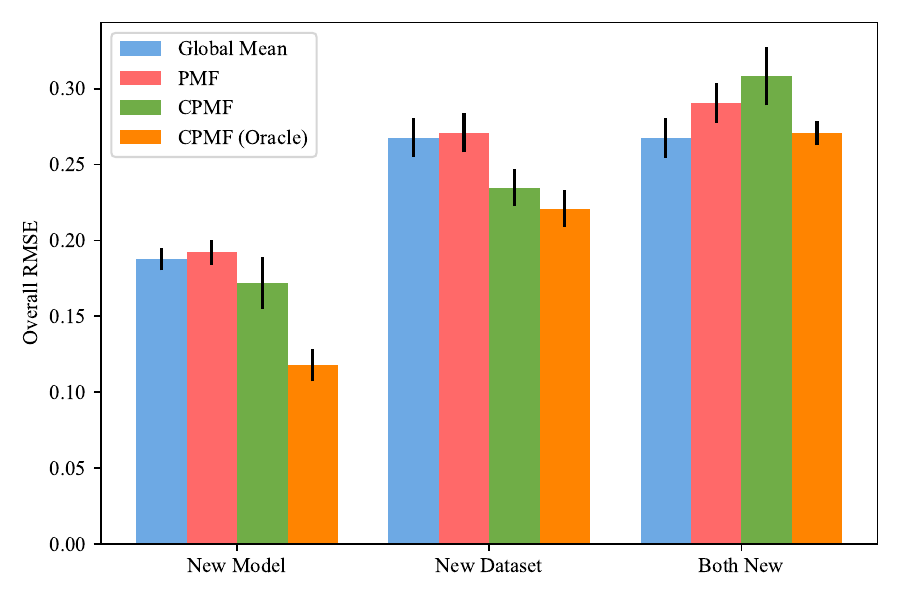}
  \caption{\textbf{Results on Purely New Models and Datasets.}} 
  \label{fig:purely_new}
\end{figure*}

We also investigate the models' ability to generalize to new models and datasets without any performance scores for training. As illustrated in \figref{fig:purely_new}, using model and dataset profiles provides slight improvement for new models or datasets. However, when both the model and dataset are entirely new, performance falls below the Global Mean baseline. But we argue that this situation is rare in practice. Some initial performance scores are usually available when a model or dataset is released, and the community usually reports more performance scores in subsequent works.

\clearpage

\begin{table*}[t]
\centering
\scriptsize
\caption{\textbf{Evaluation Cost} in different settings.}
\setlength{\tabcolsep}{1.2mm}{
\begin{threeparttable}
\begin{tabular}{lcccccc}
\toprule
Setting & Basic & vllm & LMMs-Lite & Larger model & Video benchmarks & Test-time scaling \\
\midrule
Use vllm & N & Y & Y & N (no memory) & Y & Y \\
Model Size & 7B & 7B & 7B & 32B & 7B & 7B \\
Dataset & MMBench En & MMBench En & MMBench En Lite & MMBench En & VideoMME & MathVision \\
Num of Samples & 6.72K & 6.72K & 500 & 6.72K & 2.7K & 5.14K \\
Real Performance? & Y & Possibly lower & Estimated & Y & Y & Y \\
Time & 24min & 13min & 1min & 45min & $>$2h\tnote{*} & 1h \\
Time per sample & 0.21s & 0.11s & 0.11s & 0.40s & $>$1s & 0.70s \\
\bottomrule
\end{tabular}
\begin{tablenotes}
\item[*] Video loading and preprocessing are the botteneck.
\end{tablenotes}
\end{threeparttable}
}\label{tab:eval_efficiency}
\end{table*}

\begin{table*}
\tiny 
  \centering
  \caption{\textbf{Dataset Information.} Our study utilizes 36 benchmarks. For larger benchmarks such as SEED-2, we divide them into sub-datasets based on task categories. To reduce computational costs, we subsample certain benchmarks.}
\setlength{\tabcolsep}{0.6mm}{
    \begin{tabular}{lcccl}
    \toprule
    Benchmark & \makecell{No. of \\Datasets} & \makecell{No. of Samples \\ for GPT and Gemini} & \makecell{No. of Samples \\ for Other Models} & Download URL\\
    \midrule
    \rowcolor{gray! 20} SEED 2~\citep{seed2} & 27 & 2606 & 24371 & https://huggingface.co/datasets/lmms-lab/SEED-Bench-2 \\
MME~\citep{fu2023mme} & 14 & 1000 & 2374 & https://huggingface.co/datasets/lmms-lab/MME \\
\rowcolor{gray! 20} MMBench CN~\citep{mmbench} & 20 & 1994 & 4329 & https://huggingface.co/datasets/lmms-lab/MMBench \\
MMBench EN~\citep{mmbench} & 20 & 1994 & 4329 & https://huggingface.co/datasets/lmms-lab/MMBench \\
\rowcolor{gray! 20} MMMU~\citep{mmmu} & 30 & 900 & 900 & https://huggingface.co/datasets/lmms-lab/MMMU \\
CMMMU~\citep{cmmmu} & 6 & 573 & 900 & https://huggingface.co/datasets/lmms-lab/CMMMU \\
\rowcolor{gray! 20} ScienceQA~\citep{scienceqa} & 25 & 1467 & 2017 & https://huggingface.co/datasets/lmms-lab/ScienceQA \\
CVBench~\citep{cambrian} & 4 & 400 & 2638 & https://huggingface.co/datasets/nyu-visionx/CV-Bench \\
\rowcolor{gray! 20} POPE~\citep{pope_benchmark} & 3 & 900 & 900 & https://github.com/AoiDragon/POPE \\
DECIMER~\citep{decimer} & 1 & 100 & 100 & https://www.kaggle.com/datasets/juliajakubowska/decimer \\
\rowcolor{gray! 20} Enrico~\citep{enrico} & 1 & 100 & 100 & https://userinterfaces.aalto.fi/enrico/ \\
FaceEmotion~\citep{goodfellow2013challenges} & 1 & 100 & 100 & https://www.kaggle.com/datasets/msambare/fer2013 \\
\rowcolor{gray! 20} Flickr30k~\citep{flickr30k} & 1 & 100 & 100 & https://www.kaggle.com/datasets/hsankesara/flickr-image-dataset \\
GQA~\citep{gqa} & 1 & 100 & 100 & https://cs.stanford.edu/people/dorarad/gqa/download.html \\
\rowcolor{gray! 20} HatefulMemes~\citep{kiela2020hateful} & 1 & 100 & 100 & https://www.kaggle.com/datasets/parthplc/facebook-hateful-meme-dataset \\
INAT~\citep{inat} & 1 & 100 & 100 & https://ml-inat-competition-datasets.s3.amazonaws.com/2021/val.tar.gz \\
\rowcolor{gray! 20} IRFL~\citep{irfl} & 1 & 100 & 100 & https://huggingface.co/datasets/lampent/IRFL \\
MemeCaps~\citep{hwang2023memecap} & 1 & 100 & 100 & https://github.com/eujhwang/meme-cap/tree/main \\
\rowcolor{gray! 20} Memotion~\citep{chhavi2020memotion} & 1 & 100 & 100 & https://www.kaggle.com/datasets/williamscott701/memotion-dataset-7k \\
MMIMDB~\citep{MMIMDB} & 1 & 100 & 100 & https://huggingface.co/datasets/akshayg08/mmimdb\_test \\
\rowcolor{gray! 20} NewYorkerCartoon~\citep{nycap} & 1 & 100 & 100 & https://github.com/nextml/caption-contest-data \\
NLVR~\citep{nlvr} & 1 & 100 & 100 & https://github.com/lil-lab/nlvr.git \\
\rowcolor{gray! 20} NLVR2~\citep{nlvr2} & 1 & 100 & 100 & https://github.com/lil-lab/nlvr.git \\
NoCaps~\citep{nocaps} & 1 & 100 & 100 & https://huggingface.co/datasets/akshayg08/NocapsTest \\
\rowcolor{gray! 20} OKVQA~\citep{okvqa} & 1 & 100 & 100 & https://okvqa.allenai.org/download.html \\
OpenPath~\citep{openpath} & 1 & 100 & 100 & https://huggingface.co/datasets/akshayg08/OpenPath \\
\rowcolor{gray! 20} PathVQA~\citep{pathvqa} & 1 & 100 & 100 & https://github.com/UCSD-AI4H/PathVQA \\
Resisc45~\citep{Resisc45} & 1 & 100 & 100 & https://www.kaggle.com/datasets/happyyang/nwpu-data-set \\
\rowcolor{gray! 20} Screen2Words~\citep{screen2words} & 1 & 100 & 100 & https://www.kaggle.com/datasets/onurgunes1993/rico-dataset \\
Slake~\citep{slake} & 1 & 100 & 100 & https://huggingface.co/datasets/BoKelvin/SLAKE/ \\
\rowcolor{gray! 20} UCMerced~\citep{UCMerced} & 1 & 100 & 100 & https://www.kaggle.com/code/apollo2506/land-scene-classification \\
VCR~\citep{zellers2019vcr} & 1 & 100 & 100 & https://visualcommonsense.com/download/ \\
\rowcolor{gray! 20} VisualGenome~\citep{vg} & 1 & 100 & 100 & https://homes.cs.washington.edu/~ranjay/visualgenome/ \\
VQA~\citep{antol2015vqa} & 1 & 100 & 100 & https://visualqa.org/vqa\_v1\_download.html \\
\rowcolor{gray! 20} VQARAD~\citep{vqarad} & 1 & 100 & 100 & https://huggingface.co/datasets/flaviagiammarino/vqa-rad \\
Winoground~\citep{winoground} & 1 & 100 & 100 & https://huggingface.co/datasets/facebook/winoground \\

\rowcolor{gray! 20} MathVision~\citep{mathvision} & 1 & Not evaluated & 3040 & https://huggingface.co/datasets/MathLLMs/MathVision \\
EMMA~\citep{emma} & 4 & Not evaluated & 2788 & https://huggingface.co/datasets/luckychao/EMMA \\
\rowcolor{gray! 20} Video-MME~\citep{videomme} & 1 & Not evaluated & 2700 & https://huggingface.co/datasets/lmms-lab/Video-MME \\
LongVideoBench~\citep{longvideobench} & 1 & Not evaluated & 1337 & https://huggingface.co/datasets/longvideobench/LongVideoBench \\
    \bottomrule
    \end{tabular}}
  \label{tab:dataset_info}%
\end{table*}

\begin{table*}
\tiny 
  \centering
  \caption{\textbf{Dataset Metrics.} PMF models the main metric on the datasets, while PTF utilizes the main and other metrics (six in total) in modeling. BARTScore is proposed by \citet{bartscore}, while BERTScore is introduced by \citet{bertscore}.}
\setlength{\tabcolsep}{1.5mm}{
    \begin{tabular}{lcc}
    \toprule
    Benchmark & Main Metric & Other Metrics \\
    \midrule
    \rowcolor{gray! 20} SEED 2~\citep{seed2} & Accuracy & - \\
MME~\citep{fu2023mme} & Accuracy & Precision, Recall, F1 \\
\rowcolor{gray! 20} MMBench CN~\citep{mmbench} & Accuracy & - \\
MMBench EN~\citep{mmbench} & Accuracy & - \\
\rowcolor{gray! 20} MMMU~\citep{mmmu} & Accuracy & - \\
CMMMU~\citep{cmmmu} & Accuracy & - \\
\rowcolor{gray! 20} ScienceQA~\citep{scienceqa} & Accuracy & - \\
CVBench~\citep{cambrian} & Accuracy & - \\
\rowcolor{gray! 20} POPE~\citep{pope_benchmark} & Accuracy & Precision, Recall, F1 \\
DECIMER~\citep{decimer} & BARTScore & BERTScore \\
\rowcolor{gray! 20} Enrico~\citep{enrico} & BARTScore & BERTScore \\
FaceEmotion~\citep{goodfellow2013challenges} & BARTScore & BERTScore \\
\rowcolor{gray! 20} Flickr30k~\citep{flickr30k} & BARTScore & BERTScore \\
GQA~\citep{gqa} & BARTScore & BERTScore \\
\rowcolor{gray! 20} HatefulMemes~\citep{kiela2020hateful} & BARTScore & BERTScore \\
INAT~\citep{inat} & BARTScore & BERTScore \\
\rowcolor{gray! 20} IRFL~\citep{irfl} & BARTScore & BERTScore \\
MemeCaps~\citep{hwang2023memecap} & BARTScore & BERTScore \\
\rowcolor{gray! 20} Memotion~\citep{chhavi2020memotion} & BARTScore & BERTScore \\
MMIMDB~\citep{MMIMDB} & BARTScore & BERTScore \\
\rowcolor{gray! 20} NewYorkerCartoon~\citep{nycap} & BARTScore & BERTScore \\
NLVR~\citep{nlvr} & BARTScore & BERTScore \\
\rowcolor{gray! 20} NLVR2~\citep{nlvr2} & BARTScore & BERTScore \\
NoCaps~\citep{nocaps} & BARTScore & BERTScore \\
\rowcolor{gray! 20} OKVQA~\citep{okvqa} & BARTScore & BERTScore \\
OpenPath~\citep{openpath} & BARTScore & BERTScore \\
\rowcolor{gray! 20} PathVQA~\citep{pathvqa} & BARTScore & BERTScore \\
Resisc45~\citep{Resisc45} & BARTScore & BERTScore \\
\rowcolor{gray! 20} Screen2Words~\citep{screen2words} & BARTScore & BERTScore \\
Slake~\citep{slake} & BARTScore & BERTScore \\
\rowcolor{gray! 20} UCMerced~\citep{UCMerced} & BARTScore & BERTScore \\
VCR~\citep{zellers2019vcr} & BARTScore & BERTScore \\
\rowcolor{gray! 20} VisualGenome~\citep{vg} & BARTScore & BERTScore \\
VQA~\citep{antol2015vqa} & BARTScore & BERTScore \\
\rowcolor{gray! 20} VQARAD~\citep{vqarad} & BARTScore & BERTScore \\
Winoground~\citep{winoground} & BARTScore & BERTScore \\
\rowcolor{gray! 20} MathVision~\citep{mathvision} & Accuracy & -  \\
EMMA~\citep{emma} & Accuracy & -  \\
\rowcolor{gray! 20} Video-MME~\citep{videomme} & Accuracy & -  \\
LongVideoBench~\citep{longvideobench} & Accuracy & -  \\
    \bottomrule
    \end{tabular}}
  \label{tab:dataset_metric}%
\end{table*}

\begin{table*}
\tiny 
  \centering
  \caption{\textbf{Model Information.} Our study evaluates 108 models. For each model, we report the number of parameters in the LLM backbone, the vision encoder, and the model family that we define.}
\setlength{\tabcolsep}{1.0mm}{
    \begin{tabular}{llccc}
    \toprule
    Model & Checkpoint & No. Param. in LLM (B) & Vision Encoder & Model Family \\
    \rowcolor{gray!20} BLIP2 & BLIP2-opt-2.7B & 2.7 & ViT & BLIP\\
  & BLIP2-flan-t5-xxl & 11 & ViT & BLIP\\
\rowcolor{gray!20}   & BLIP2-opt-6.7b-coco & 6.7 & ViT & BLIP\\
  & BLIP2-opt-6.7b & 6.7 & ViT & BLIP\\
\rowcolor{gray!20}   & BLIP2-flan-t5-xl & 3 & ViT & BLIP\\
InstructBLIP & InstructBLIP-Vicuna-7B & 7 & ViT & BLIP\\
\rowcolor{gray!20}   & InstructBLIP-Vicuna-13B & 13 & ViT & BLIP\\
  & InstructBLIP-flan-t5-xl & 3 & ViT & BLIP\\
\rowcolor{gray!20}   & InstructBLIP-flan-t5-xxl & 11 & ViT & BLIP\\
MiniGPT4 & MiniGPT4-LLaMA2-7B & 7 & ViT & MiniGPT4\\
\rowcolor{gray!20}   & MiniGPT4-Vicuna0-7B & 7 & ViT & MiniGPT4\\
  & MiniGPT4-Vicuna0-13B & 13 & ViT & MiniGPT4\\
\rowcolor{gray!20} mPLUG-Owl & mPLUG-Owl2-LLaMA2-7B & 7 & ViT & MiniGPT4\\
  & mPLUG-Owl2\_1 & 7 & ViT & mPLUG-Owl\\
\rowcolor{gray!20} LLaVA & LLaVA-7B & 7 & CLIP & LLaVA\\
  & LLaVA-13B & 13 & CLIP & LLaVA\\
\rowcolor{gray!20}   & LLaVA-v1.6-Vicuna-7B & 7 & CLIP & LLaVA\\
  & LLaVA-v1.6-Vicuna-13B & 13 & CLIP & LLaVA\\
\rowcolor{gray!20}   & LLaVA-v1.6-Mistral-7B & 7 & CLIP & LLaVA\\
  & LLaVA-v1.6-34B & 34 & CLIP & LLaVA\\
\rowcolor{gray!20} Cambrian-1 & Cambrian-Phi3-3B & 3 & CLIP, SigLIP, ConvNeXt, DINOv2 & Cambrian\\
  & Cambrian-8B & 8 & CLIP, SigLIP, ConvNeXt, DINOv2 & Cambrian\\
\rowcolor{gray!20}   & Cambrian-13B & 13 & CLIP, SigLIP, ConvNeXt, DINOv2 & Cambrian\\
  & Cambrian-34B & 34 & CLIP, SigLIP, ConvNeXt, DINOv2 & Cambrian\\
\rowcolor{gray!20} Fuyu & Fuyu-8B & 8 & - & Fuyu\\
LLaMA\_Adapter & LLaMA-Adapter-V2-BIAS-7B & 7 & CLIP & LLaMA-Adapter\\
\rowcolor{gray!20}   & LLaMA-Adapter-V2-LORA-BIAS-7B & 7 & CLIP & LLaMA-Adapter\\
  & LLaMA-Adapter-V2-LORA-BIAS-7B-v21 & 7 & CLIP & LLaMA-Adapter\\
\rowcolor{gray!20} OpenFlamingo & OpenFlamingo-3B-vitl-mpt1b & 1 & NFNet & OpenFlamingo\\
  & OpenFlamingo-3B-vitl-mpt1b-langinstruct & 1 & NFNet & OpenFlamingo\\
\rowcolor{gray!20}   & OpenFlamingo-4B-vitl-rpj3b & 3 & NFNet & OpenFlamingo\\
  & OpenFlamingo-4B-vitl-rpj3b-langinstruct & 3 & NFNet & OpenFlamingo\\
\rowcolor{gray!20}   & OpenFlamingo-9B-vitl-mpt7b & 7 & NFNet & OpenFlamingo\\
Qwen-VL & Qwen-VL-Chat & 7 & ViT & Qwen\\
\rowcolor{gray!20} & Qwen2-VL-2B-Instruct & 2 & ViT & Qwen \\
& Qwen2-VL-7B-Instruct & 7 & ViT & Qwen \\
\rowcolor{gray!20} & Qwen2.5-VL-3B-Instruct & 3 & ViT & Qwen \\
& Qwen2.5-VL-7B-Instruct & 7 & ViT & Qwen \\
\rowcolor{gray!20} & Qwen2.5-VL-32B-Instruct & 32 & ViT & Qwen \\

DeepSeek-VL & DeepSeek-VL2-tiny & 10 & CLIP & Qwen \\
\rowcolor{gray!20} & DeepSeek-VL2-small & 28 & CLIP & Qwen \\

InternLM\_XComposer & InternLM-XComposer-7B & 7 & CLIP & InternLM\\
\rowcolor{gray!20}   & InternLM-XComposer-vl-7B & 7 & CLIP & InternLM\\
 & InternLM-XComposer2-7B & 7 & CLIP & InternLM\\
\rowcolor{gray!20}   & InternLM-XComposer2-vl-1\_8b & 1.8 & CLIP & InternLM\\
  & InternLM-XComposer2-vl-7B & 7 & CLIP & InternLM\\
\rowcolor{gray!20} GPT4 & gpt-4o-2024-05-13 & Unknown & Unknown & GPT4\\
 & gpt-4o-2024-08-06 & Unknown & Unknown & GPT4\\
\rowcolor{gray!20}   & gpt-4o-mini-2024-07-18 & Unknown & Unknown & GPT4\\
 & gpt-4-turbo-2024-04-09 & Unknown & Unknown & GPT4\\
\rowcolor{gray!20} Gemini & gemini-1.5-pro & Unknown & Unknown & Gemini\\
 & gemini-1.5-flash & Unknown & Unknown & Gemini\\
    \bottomrule
    \end{tabular}}
  \label{tab:model_info}%
\end{table*}

\begin{table*}
\tiny 
  \centering
  \caption{\textbf{Model Information.} This is the continued table of \tabref{tab:model_info}}
\setlength{\tabcolsep}{1.0mm}{
    \begin{tabular}{llccc}
    \toprule
    Model & Checkpoint & No. Param. in LLM & Vision Encoder & Model Family \\
Prismatic & reproduction-llava-v15+7b & 7 & CLIP & prism\\
\rowcolor{gray!20}   & reproduction-llava-v15+13b & 13 & CLIP & prism\\
  & one-stage+7b & 7 & CLIP & prism\\
\rowcolor{gray!20}   & one-stage+13b & 13 & CLIP & prism\\
  & full-ft-multi-stage+7b & 7 & CLIP & prism\\
\rowcolor{gray!20}   & full-ft-one-stage+7b & 7 & CLIP & prism\\
  & in1k-224px+7b & 7 & ViT & prism\\
\rowcolor{gray!20}   & dinov2-224px+7b & 7 & DINOv2 & prism\\
  & clip-224px+7b & 7 & CLIP & prism\\
\rowcolor{gray!20}   & siglip-224px+7b & 7 & SigLIP & prism\\
  & clip-336px-resize-crop+7b & 7 & CLIP & prism\\
\rowcolor{gray!20}   & clip-336px-resize-naive+7b & 7 & CLIP & prism\\
  & siglip-384px-letterbox+7b & 7 & SigLIP & prism\\
\rowcolor{gray!20}   & siglip-384px-resize-crop+7b & 7 & SigLIP & prism\\
  & siglip-384px-resize-naive+7b & 7 & SigLIP & prism\\
\rowcolor{gray!20}   & dinoclip-336px-letterbox+7b & 7 & CLIP, DINOv2 & prism\\
  & dinoclip-336px-resize-naive+7b & 7 & CLIP, DINOv2 & prism\\
\rowcolor{gray!20}   & dinosiglip-384px-letterbox+7b & 7 & SigLIP, DINOv2 & prism\\
  & dinosiglip-384px-resize-naive+7b & 7 & SigLIP, DINOv2 & prism\\
\rowcolor{gray!20}   & llama2+7b & 7 & CLIP & prism\\
  & llama2+13b & 13 & CLIP & prism\\
\rowcolor{gray!20}   & vicuna-no-cotraining+7b & 7 & CLIP & prism\\
  & llama2-no-cotraining+7b & 7 & CLIP & prism\\
\rowcolor{gray!20}   & train-1.25-epochs+7b & 7 & CLIP & prism\\
  & train-1.5-epochs+7b & 7 & CLIP & prism\\
\rowcolor{gray!20}   & train-2-epochs+7b & 7 & CLIP & prism\\
  & train-3-epochs+7b & 7 & CLIP & prism\\
\rowcolor{gray!20}   & llava-lvis4v+7b & 7 & CLIP & prism\\
  & llava-lrv+7b & 7 & CLIP & prism\\
\rowcolor{gray!20}   & llava-lvis4v-lrv+7b & 7 & CLIP & prism\\
  & prism-clip-controlled+7b & 7 & CLIP & prism\\
\rowcolor{gray!20}   & prism-clip-controlled+13b & 13 & CLIP & prism\\
  & prism-clip+7b & 7 & CLIP & prism\\
\rowcolor{gray!20}   & prism-clip+13b & 13 & CLIP & prism\\
  & prism-siglip-controlled+7b & 7 & SigLIP & prism\\
\rowcolor{gray!20}   & prism-siglip-controlled+13b & 13 & SigLIP & prism\\
  & prism-siglip+7b & 7 & SigLIP & prism\\
\rowcolor{gray!20}   & prism-siglip+13b & 13 & SigLIP & prism\\
  & prism-dinosiglip-controlled+7b & 7 & SigLIP, DINOv2 & prism\\
\rowcolor{gray!20}   & prism-dinosiglip-controlled+13b & 13 & SigLIP, DINOv2 & prism\\
  & prism-dinosiglip+7b & 7 & SigLIP, DINOv2 & prism\\
\rowcolor{gray!20}   & prism-dinosiglip+13b & 13 & SigLIP, DINOv2 & prism\\
  & prism-dinosiglip-224px-controlled+7b & 7 & SigLIP, DINOv2 & prism\\
\rowcolor{gray!20}   & prism-dinosiglip-224px+7b & 7 & SigLIP, DINOv2 & prism\\
  & llama2-chat+13b & 13 & CLIP & prism\\
\rowcolor{gray!20}   & mistral-v0.1+7b & 7 & CLIP & prism\\
  & mistral-instruct-v0.1+7b & 7 & CLIP & prism\\
\rowcolor{gray!20}   & phi-2+3b & 3 & CLIP & prism\\
  & gemma-instruct+2b+clip & 2 & CLIP & prism\\
\rowcolor{gray!20}   & gemma-instruct+2b+siglip & 2 & SigLIP & prism\\
  & gemma-instruct+2b+dinosiglip & 2 & SigLIP, DINOv2 & prism\\
\rowcolor{gray!20}   & gemma-instruct+8b+clip & 8 & CLIP & prism\\
  & gemma-instruct+8b+siglip & 8 & SigLIP & prism\\
\rowcolor{gray!20}   & gemma-instruct+8b+dinosiglip & 8 & SigLIP, DINOv2 & prism\\
  & llama2-chat+7b+clip & 7 & CLIP & prism\\
\rowcolor{gray!20}   & llama2-chat+7b+siglip & 7 & SigLIP & prism\\
  & llama2-chat+7b+dinosiglip & 7 & SigLIP, DINOv2 & prism\\
\rowcolor{gray!20}   & llama3-instruct+8b+clip & 8 & CLIP & prism\\
  & llama3-instruct+8b+siglip & 8 & SigLIP & prism\\
\rowcolor{gray!20}   & llama3-instruct+8b+dinosiglip & 8 & SigLIP, DINOv2 & prism\\
  & mistral-instruct-v0.2+7b+clip & 7 & CLIP & prism\\
\rowcolor{gray!20}   & mistral-instruct-v0.2+7b+siglip & 7 & SigLIP & prism\\
  & mistral-instruct-v0.2+7b+dinosiglip & 7 & SigLIP, DINOv2 & prism\\
    \bottomrule
    \end{tabular}}
  \label{tab:model_info_2}%
\end{table*}

\begin{table*}[htbp]
\scriptsize
  \centering
  \caption{\textbf{Comparison of PMF and Deep Learning-Based Methods. }}
    \begin{tabular}{lcccccc}
    \toprule
    \multirow{2}[4]{*}{Method} & \multicolumn{2}{c}{Overall} & \multicolumn{2}{c}{Acc} & \multicolumn{2}{c}{BART} \\
    \cmidrule{2-7}
    & RMSE$\downarrow$ & MAE$\downarrow$ & RMSE & MAE & RMSE & MAE \\
    \midrule
    \multicolumn{7}{l}{\textit{Test ratio: 20\%}} \\
    DMF [1] & 0.225 & 0.105 & 0.086 & 0.060 & 0.538 & 0.353 \\
    GCMC [2] &  &  & 0.187 & 0.139 &  &  \\
    PMF & 0.193 & 0.090 & 0.074 & 0.052 & 0.461 & 0.303 \\
    \midrule
    \multicolumn{7}{l}{\textit{Test ratio: 80\%}} \\
    DMF [1] & 0.561 & 0.314 & 0.289 & 0.209 & 1.26 & 0.896 \\
    GCMC [2] &  &  & - & - &  &  \\
    PMF & 0.317 & 0.174 & 0.156 & 0.115 & 0.723 & 0.504 \\
    \bottomrule
    \end{tabular}%
  \label{tab:pmf_vs_dmf}%
\end{table*}%

\begin{table*}
\scriptsize 
  \centering
  \caption{\textbf{Detailed Performance of PMF.} Superior results are highlighted.}
\setlength{\tabcolsep}{1.2mm}{
    \begin{tabular}{lcccccccc}
    \toprule
    \multirow{2}[4]{*}{Method} & \multicolumn{2}{c}{Overall} & \multicolumn{3}{c}{Acc}
    & \multicolumn{3}{c}{BART} \\ 
    \cmidrule{2-9}
    & RMSE$\downarrow$ & MAE$\downarrow$ & RMSE & MAE & $R^2\uparrow$ & RMSE & MAE & $R^2$ \\
    \midrule
\multicolumn{2}{l}{Test Ratio: 20\%}\\
Global Mean & 0.367 & 0.220 & 0.190 & 0.148 & 0.475 & 0.823 & 0.611 & 0.549\\
Mean Of Means & 0.319 & 0.186 & 0.161 & 0.125 & 0.622 & 0.719 & 0.525 & 0.656\\
PMF & \cellcolor{orange!20}0.192 & \cellcolor{orange!20}0.090 & \cellcolor{orange!20}0.073 & \cellcolor{orange!20}0.051 & \cellcolor{orange!20} 0.922 & \cellcolor{orange!20}0.458 & \cellcolor{orange!20}0.303 & \cellcolor{orange!20} 0.860\\
\cmidrule{1-9}
\multicolumn{2}{l}{Test Ratio: 40\%}\\
Global Mean & 0.368 & 0.220 & 0.190 & 0.149 & 0.477 & 0.829 & 0.614 & 0.551\\
Mean Of Means & 0.320 & 0.186 & 0.161 & 0.125 & 0.623 & 0.725 & 0.527 & 0.657\\
PMF & \cellcolor{orange!20}0.199 & \cellcolor{orange!20}0.095 & \cellcolor{orange!20}0.078 & \cellcolor{orange!20}0.056 & \cellcolor{orange!20} 0.911 & \cellcolor{orange!20}0.474 & \cellcolor{orange!20}0.314 & \cellcolor{orange!20} 0.853\\
\cmidrule{1-9}
\multicolumn{2}{l}{Test Ratio: 60\%}\\
Global Mean & 0.370 & 0.220 & 0.190 & 0.149 & 0.474 & 0.831 & 0.613 & 0.551\\
Mean Of Means & 0.322 & 0.188 & 0.162 & 0.126 & 0.618 & 0.729 & 0.529 & 0.654\\
PMF & \cellcolor{orange!20}0.220 & \cellcolor{orange!20}0.106 & \cellcolor{orange!20}0.089 & \cellcolor{orange!20}0.063 & \cellcolor{orange!20} 0.886 & \cellcolor{orange!20}0.521 & \cellcolor{orange!20}0.348 & \cellcolor{orange!20} 0.823\\
\cmidrule{1-9}
\multicolumn{2}{l}{Test Ratio: 80\%}\\
Global Mean & 0.373 & 0.221 & 0.193 & 0.150 & 0.462 & 0.837 & 0.612 & 0.546\\
Mean Of Means & 0.329 & 0.191 & 0.166 & 0.128 & 0.601 & 0.742 & 0.533 & 0.643\\
PMF & \cellcolor{orange!20}0.258 & \cellcolor{orange!20}0.131 & \cellcolor{orange!20}0.114 & \cellcolor{orange!20}0.081 & \cellcolor{orange!20} 0.812 & \cellcolor{orange!20}0.600 & \cellcolor{orange!20}0.407 & \cellcolor{orange!20} 0.766\\
\cmidrule{1-9}
\multicolumn{2}{l}{Test Ratio: 90\%}\\
Global Mean & 0.381 & 0.226 & 0.198 & 0.153 & 0.430 & 0.852 & 0.630 & 0.529\\
Mean Of Means & 0.339 & 0.197 & 0.174 & 0.133 & 0.564 & 0.765 & 0.555 & 0.621\\
PMF & \cellcolor{orange!20}0.313 & \cellcolor{orange!20}0.172 & \cellcolor{orange!20}0.153 & \cellcolor{orange!20}0.113 & \cellcolor{orange!20} 0.660 & \cellcolor{orange!20}0.714 & \cellcolor{orange!20}0.502 & \cellcolor{orange!20} 0.669\\
\cmidrule{1-9}
\multicolumn{2}{l}{Test Ratio: 95\%}\\
Global Mean & 0.458 & 0.278 & 0.227 & 0.182 & 0.254 & 1.041 & 0.807 & 0.297\\
Mean Of Means & \cellcolor{orange!20}0.385 & \cellcolor{orange!20}0.230 & \cellcolor{orange!20}0.191 & \cellcolor{orange!20}0.150 & \cellcolor{orange!20} 0.474 & \cellcolor{orange!20}0.875 & \cellcolor{orange!20}0.672 & \cellcolor{orange!20} 0.504\\
PMF & 0.462 & 0.276 & 0.228 & 0.180 & 0.248 & 1.052 & 0.805 & 0.282\\
    \bottomrule
    \end{tabular}}
  \label{tab:detailed_pmf}%
\end{table*}%

\begin{table*}
\scriptsize 
  \centering
  \caption{\textbf{Detailed Performance of PTF.} Superior results are highlighted.}
\setlength{\tabcolsep}{1.2mm}{
    \begin{tabular}{lcccccccccccccc}
    \toprule
    \multirow{2}[4]{*}{Method} & \multicolumn{2}{c}{Overall} & \multicolumn{2}{c}{Acc} & \multicolumn{2}{c}{Precision}
    & \multicolumn{2}{c}{Recall}
    & \multicolumn{2}{c}{F1}
    & \multicolumn{2}{c}{BART}
    & \multicolumn{2}{c}{BERT} \\ 
    \cmidrule{2-15}
    & RMSE$\downarrow$ & MAE$\downarrow$ & RMSE & MAE & RMSE & MAE & RMSE & MAE & RMSE & MAE & RMSE & MAE & RMSE & MAE \\
    \multicolumn{2}{l}{Test Ratio: 20\%}\\
Global Mean & 0.320 & 0.190 & 0.190 & 0.149 & 0.223 & 0.167 & 0.245 & 0.186 & 0.205 & 0.156 & 0.812 & 0.603 & 0.088 & 0.044\\
Mean Of Means & 0.260 & 0.150 & 0.149 & 0.115 & 0.181 & 0.131 & 0.205 & 0.152 & 0.169 & 0.123 & 0.664 & 0.482 & \cellcolor{orange!20} 0.074 & \cellcolor{orange!20} 0.036\\
PMF & 0.206 & 0.096 & 0.081 & 0.057 & 0.130 & 0.086 & 0.175 & 0.126 & 0.108 & \cellcolor{orange!20} 0.070 & 0.563 & 0.378 & 0.077 & 0.039\\
CPTF & 0.208 & 0.099 & 0.087 & 0.060 & 0.134 & 0.089 & \cellcolor{orange!20} 0.173 & \cellcolor{orange!20} 0.123 & \cellcolor{orange!20} 0.107 & 0.070 & 0.564 & 0.379 & 0.076 & 0.039\\
BPTF & \cellcolor{orange!20} 0.202 & \cellcolor{orange!20} 0.095 & 0.079 & 0.056 & \cellcolor{orange!20} 0.129 & \cellcolor{orange!20} 0.084 & 0.177 & 0.127 & 0.109 & 0.070 & \cellcolor{orange!20} 0.553 & \cellcolor{orange!20} 0.372 & 0.077 & 0.039\\
BCPTF & 0.207 & 0.096 & \cellcolor{orange!20} 0.079 & \cellcolor{orange!20} 0.056 & 0.129 & 0.085 & 0.178 & 0.127 & 0.113 & 0.072 & 0.568 & 0.378 & 0.076 & 0.039\\
\cmidrule{1-15}
\multicolumn{2}{l}{Test Ratio: 40\%}\\
Global Mean & 0.323 & 0.192 & 0.190 & 0.149 & 0.223 & 0.167 & 0.247 & 0.189 & 0.213 & 0.160 & 0.818 & 0.609 & 0.091 & 0.045\\
Mean Of Means & 0.262 & 0.151 & 0.150 & 0.116 & 0.182 & 0.132 & 0.209 & 0.156 & 0.174 & 0.126 & 0.667 & 0.486 & \cellcolor{orange!20} 0.078 & \cellcolor{orange!20} 0.038\\
PMF & 0.210 & 0.100 & 0.083 & 0.059 & 0.132 & 0.087 & 0.181 & 0.130 & \cellcolor{orange!20} 0.112 & \cellcolor{orange!20} 0.073 & 0.572 & 0.388 & 0.081 & 0.040\\
CPTF & 0.209 & 0.101 & 0.087 & 0.062 & 0.134 & 0.089 & \cellcolor{orange!20} 0.180 & \cellcolor{orange!20} 0.128 & 0.113 & 0.074 & 0.566 & 0.385 & 0.081 & 0.040\\
BPTF & \cellcolor{orange!20} 0.206 & 0.100 & 0.084 & 0.060 & 0.132 & 0.086 & 0.185 & 0.133 & 0.117 & 0.077 & \cellcolor{orange!20} 0.558 & \cellcolor{orange!20} 0.379 & 0.082 & 0.041\\
BCPTF & 0.209 & \cellcolor{orange!20} 0.100 & \cellcolor{orange!20} 0.082 & \cellcolor{orange!20} 0.059 & \cellcolor{orange!20} 0.131 & \cellcolor{orange!20} 0.086 & 0.184 & 0.131 & 0.117 & 0.076 & 0.568 & 0.384 & 0.081 & 0.040\\
\cmidrule{1-15}
\multicolumn{2}{l}{Test Ratio: 60\%}\\
Global Mean & 0.325 & 0.192 & 0.191 & 0.149 & 0.227 & 0.170 & 0.248 & 0.189 & 0.214 & 0.160 & 0.825 & 0.611 & 0.092 & 0.045\\
Mean Of Means & 0.265 & 0.153 & 0.151 & 0.116 & 0.186 & 0.135 & 0.212 & 0.157 & 0.178 & 0.128 & 0.676 & 0.490 & \cellcolor{orange!20} 0.080 & \cellcolor{orange!20} 0.038\\
PMF & 0.218 & 0.107 & 0.093 & 0.067 & 0.136 & 0.090 & 0.188 & 0.134 & \cellcolor{orange!20} 0.123 & \cellcolor{orange!20} 0.081 & 0.588 & 0.400 & 0.084 & 0.041\\
CPTF & 0.219 & 0.109 & 0.098 & 0.070 & 0.141 & 0.094 & \cellcolor{orange!20} 0.187 & \cellcolor{orange!20} 0.133 & 0.125 & 0.082 & 0.588 & 0.398 & 0.083 & 0.040\\
BPTF & 0.217 & 0.108 & 0.096 & 0.068 & 0.138 & 0.092 & 0.194 & 0.139 & 0.130 & 0.087 & \cellcolor{orange!20} 0.582 & 0.397 & 0.085 & 0.042\\
BCPTF & \cellcolor{orange!20} 0.216 & \cellcolor{orange!20} 0.105 & \cellcolor{orange!20} 0.089 & \cellcolor{orange!20} 0.064 & \cellcolor{orange!20} 0.135 & \cellcolor{orange!20} 0.090 & 0.191 & 0.136 & 0.127 & 0.083 & 0.584 & \cellcolor{orange!20} 0.394 & 0.083 & 0.041\\
\cmidrule{1-15}
\multicolumn{2}{l}{Test Ratio: 80\%}\\
Global Mean & 0.330 & 0.194 & 0.193 & 0.150 & 0.230 & 0.171 & 0.253 & 0.191 & 0.217 & 0.162 & 0.839 & 0.619 & 0.092 & 0.046\\
Mean Of Means & 0.277 & 0.158 & 0.155 & 0.119 & 0.198 & 0.141 & 0.225 & 0.165 & 0.186 & 0.134 & 0.709 & 0.510 & \cellcolor{orange!20} 0.084 & \cellcolor{orange!20} 0.041\\
PMF & 0.249 & 0.128 & 0.120 & 0.087 & 0.151 & 0.103 & \cellcolor{orange!20} 0.207 & 0.148 & \cellcolor{orange!20} 0.145 & \cellcolor{orange!20} 0.098 & 0.661 & 0.457 & 0.091 & 0.044\\
CPTF & 0.240 & 0.123 & 0.115 & 0.083 & 0.151 & 0.104 & 0.208 & \cellcolor{orange!20} 0.148 & 0.147 & 0.099 & 0.637 & 0.437 & 0.088 & 0.043\\
BPTF & 0.239 & 0.123 & 0.116 & 0.083 & 0.152 & 0.103 & 0.212 & 0.151 & 0.151 & 0.102 & 0.630 & 0.433 & 0.090 & 0.044\\
BCPTF & \cellcolor{orange!20} 0.236 & \cellcolor{orange!20} 0.119 & \cellcolor{orange!20} 0.108 & \cellcolor{orange!20} 0.077 & \cellcolor{orange!20} 0.147 & \cellcolor{orange!20} 0.099 & 0.208 & 0.149 & 0.147 & 0.099 & \cellcolor{orange!20} 0.627 & \cellcolor{orange!20} 0.427 & 0.089 & 0.043\\
\cmidrule{1-15}
\multicolumn{2}{l}{Test Ratio: 90\%}\\
Global Mean & 0.338 & 0.198 & 0.199 & 0.154 & 0.237 & 0.174 & 0.258 & 0.195 & 0.224 & 0.166 & 0.858 & 0.629 & 0.095 & 0.047\\
Mean Of Means & 0.298 & 0.168 & 0.166 & 0.125 & 0.216 & 0.153 & 0.239 & 0.176 & 0.204 & 0.147 & 0.764 & 0.547 & \cellcolor{orange!20} 0.090 & \cellcolor{orange!20} 0.043\\
PMF & 0.294 & 0.161 & 0.161 & 0.119 & 0.194 & 0.135 & 0.235 & 0.171 & 0.187 & 0.131 & 0.761 & 0.535 & 0.094 & 0.045\\
CPTF & 0.274 & 0.147 & 0.145 & 0.105 & 0.190 & 0.133 & 0.233 & 0.168 & 0.184 & 0.128 & 0.710 & 0.492 & 0.092 & 0.043\\
BPTF & \cellcolor{orange!20} 0.267 & 0.143 & 0.142 & 0.103 & \cellcolor{orange!20} 0.179 & \cellcolor{orange!20} 0.123 & 0.232 & 0.167 & 0.178 & 0.122 & \cellcolor{orange!20} 0.690 & \cellcolor{orange!20} 0.480 & 0.093 & 0.044\\
BCPTF & 0.268 & \cellcolor{orange!20} 0.141 & \cellcolor{orange!20} 0.138 & \cellcolor{orange!20} 0.099 & 0.179 & 0.124 & \cellcolor{orange!20} 0.228 & \cellcolor{orange!20} 0.164 & \cellcolor{orange!20} 0.176 & \cellcolor{orange!20} 0.120 & 0.698 & 0.481 & 0.093 & 0.045\\
\cmidrule{1-15}
\multicolumn{2}{l}{Test Ratio: 95\%}\\
Global Mean & 0.404 & 0.238 & 0.228 & 0.182 & 0.251 & 0.194 & 0.270 & 0.209 & 0.240 & 0.183 & 1.058 & 0.805 & 0.101 & 0.057\\
Mean Of Means & 0.387 & 0.217 & 0.202 & 0.158 & 0.241 & 0.182 & 0.261 & 0.198 & 0.230 & 0.172 & 1.027 & 0.772 & \cellcolor{orange!20} 0.097 & 0.056\\
PMF & 0.403 & 0.233 & 0.223 & 0.177 & 0.244 & 0.185 & 0.269 & 0.204 & 0.236 & 0.175 & 1.059 & 0.801 & 0.101 & 0.057\\
CPTF & 0.364 & 0.199 & 0.188 & 0.142 & 0.216 & 0.161 & 0.252 & 0.188 & 0.216 & 0.157 & 0.970 & 0.712 & 0.097 & \cellcolor{orange!20} 0.054\\
BPTF & \cellcolor{orange!20} 0.355 & \cellcolor{orange!20} 0.196 & 0.189 & 0.144 & \cellcolor{orange!20} 0.208 & \cellcolor{orange!20} 0.153 & \cellcolor{orange!20} 0.251 & 0.188 & 0.206 & 0.148 & \cellcolor{orange!20} 0.940 & \cellcolor{orange!20} 0.687 & 0.098 & 0.055\\
BCPTF & 0.360 & 0.196 & \cellcolor{orange!20} 0.186 & \cellcolor{orange!20} 0.141 & 0.211 & 0.156 & 0.251 & \cellcolor{orange!20} 0.186 & \cellcolor{orange!20} 0.205 & \cellcolor{orange!20} 0.148 & 0.959 & 0.702 & 0.100 & 0.056\\
    \bottomrule
    \end{tabular}}
  \label{tab:detailed_ptf}%
\end{table*}%

\begin{table*}
\centering
\scriptsize
\caption{\textbf{Ranking-Based Metrics.} Superior results are highlighted.}
\label{tab:rank_performance}
\setlength{\tabcolsep}{1.2mm}{
\begin{tabular}{lcccccccc}
\toprule
Method & Spearman$\uparrow$ & Kendall$\uparrow$ & Precision@1$\uparrow$ & Precision@3$\uparrow$ & Precision@5$\uparrow$ & Recall@1$\uparrow$ & Recall@3$\uparrow$ & Recall@5$\uparrow$ \\
\midrule
Global Mean          & 0.86 & 0.76 & 0.75 & 0.96 & 1.00 & 0.75 & 0.79 & 0.80 \\
Mean Of Means        & 0.90 & 0.80 & 0.75 & 0.97 & 1.00 & 0.75 & 0.79 & 0.80 \\
PMF                  & 0.95 & 0.87 & 0.69 & 0.87 & 0.91 & 0.69 & 0.82 & 0.87 \\
BCPMF                & 0.95 & 0.87 & 0.69 & 0.87 & 0.90 & 0.69 & 0.81 & 0.87 \\
BCPMF (Unc 50\%)     & 0.96 & 0.89 & 0.76 & 0.92 & 0.95 & 0.76 & 0.86 & 0.90 \\
BCPMF (Unc 30\%)     & \cellcolor{orange!20}0.97 & \cellcolor{orange!20}0.91 & \cellcolor{orange!20}0.87 & \cellcolor{orange!20}0.98 & \cellcolor{orange!20}0.99 & \cellcolor{orange!20}0.87 & \cellcolor{orange!20}0.91 & \cellcolor{orange!20}0.93 \\
\midrule
Global Mean          & 0.27 & 0.19 & 0.20 & 0.48 & 0.66 & 0.20 & 0.21 & 0.21 \\
Mean Of Means        & 0.65 & 0.48 & 0.20 & 0.49 & 0.66 & 0.20 & 0.21 & 0.22 \\
PMF                  & 0.73 & 0.55 & 0.19 & 0.44 & 0.60 & 0.19 & 0.29 & 0.40 \\
BCPMF                & 0.75 & 0.57 & 0.21 & 0.47 & 0.62 & 0.21 & 0.36 & 0.46 \\
BCPMF (Unc 50\%)     & 0.75 & 0.61 & 0.40 & 0.64 & 0.75 & 0.40 & 0.50 & 0.58 \\
BCPMF (Unc 30\%)     & \cellcolor{orange!20}0.82 & \cellcolor{orange!20}0.71 & \cellcolor{orange!20}0.60 & \cellcolor{orange!20}0.78 & \cellcolor{orange!20}0.86 & \cellcolor{orange!20}0.60 & \cellcolor{orange!20}0.67 & \cellcolor{orange!20}0.73 \\
\bottomrule
\end{tabular}}
\end{table*}

\begin{table*}[tb]
\scriptsize
\centering
\caption{\textbf{Combining Coreset Methods with Our Approach} with Different Sampling Ratios.}
\setlength{\tabcolsep}{1.2mm}{
\begin{tabular}{lcccccc}
\toprule
\multirow{2}[4]{*}{Method} & \multicolumn{2}{c}{Overall} & \multicolumn{2}{c}{Acc} & \multicolumn{2}{c}{BART} \\
\cmidrule{2-7}
& RMSE$\downarrow$ & MAE$\downarrow$ & RMSE & MAE & RMSE & MAE \\
\midrule
Select 5\% Samples \\
Ours                          & 0.193 & 0.090 & 0.074 & 0.052 & 0.459 & 0.299 \\
Random Selection              & 0.345 & 0.224 & 0.250 & 0.175 & 0.652 & 0.494 \\
\quad + Ours (Avg)           & 0.199 (-0.146) & 0.126 & 0.131 & 0.093 & 0.404 & 0.306 \\
\quad + Ours (Unc)           & 0.157 (-0.188) & 0.083 & \cellcolor{orange!20}0.070 & \cellcolor{orange!20}0.050 & 0.365 & 0.261 \\
Herding                       & 0.326 & 0.220 & 0.252 & 0.177 & 0.582 & 0.458 \\
\quad + Ours (Avg)          & 0.192 (-0.134) & 0.124 & 0.133 & 0.094 & 0.377 & 0.287 \\
\quad + Ours (Unc)          & 0.155 (-0.171) & \cellcolor{orange!20}0.081 & \cellcolor{orange!20}0.070 & \cellcolor{orange!20}0.050 & 0.358 & \cellcolor{orange!20}0.252 \\
K-Center Greedy               & 0.353 & 0.231 & 0.262 & 0.182 & 0.656 & 0.498 \\
\quad + Ours (Avg)  & 0.200 (-0.153) & 0.128 & 0.137 & 0.096 & 0.394 & 0.302 \\
\quad + Ours (Unc)  & \cellcolor{orange!20}0.154 (-0.199) & 0.082 & \cellcolor{orange!20}0.070 & \cellcolor{orange!20}0.050 & \cellcolor{orange!20}0.356 & 0.258 \\
\hline
Select 10\% Samples \\
Ours                          & 0.193 & 0.090 & 0.074 & 0.052 & 0.459 & 0.299 \\
Random Selection              & 0.224 & 0.141 & 0.152 & 0.107 & 0.444 & 0.326 \\
\quad + Ours (Avg)           & 0.149 (-0.075) & 0.088 & 0.085 & 0.061 & 0.322 & 0.237 \\
\quad + Ours (Unc)           & 0.139 (-0.085) & \cellcolor{orange!20}0.076 & \cellcolor{orange!20}0.069 & \cellcolor{orange!20}0.049 & 0.313 & 0.224 \\
Herding                       & 0.216 & 0.140 & 0.155 & 0.112 & 0.410 & 0.297 \\
\quad + Ours (Avg)          & 0.144 (-0.072) & 0.088 & 0.087 & 0.064 & \cellcolor{orange!20}0.305 & \cellcolor{orange!20}0.220 \\
\quad + Ours (Unc)          & 0.140 (-0.076) & \cellcolor{orange!20}0.076 & 0.070 & \cellcolor{orange!20}0.049 & 0.315 & \cellcolor{orange!20}0.220 \\
K-Center Greedy               & 0.223 & 0.142 & 0.154 & 0.109 & 0.437 & 0.322 \\
\quad + Ours (Avg)  & 0.144 (-0.079) & 0.088 & 0.086 & 0.063 & 0.306 & 0.226 \\
\quad + Ours (Unc)  & \cellcolor{orange!20}0.138 (-0.085) & 0.077 & 0.070 & \cellcolor{orange!20}0.049 & 0.313 & 0.224 \\
\hline
Select 15\% Samples \\
Ours                          & 0.193 & 0.090 & 0.074 & 0.052 & 0.459 & 0.299 \\
Random Selection              & 0.180 & 0.114 & 0.125 & 0.087 & 0.352 & 0.261 \\
\quad + Ours (Avg)           & 0.133 (-0.047) & 0.078 & 0.073 & 0.053 & 0.291 & 0.212 \\
\quad + Ours (Unc)           & 0.132 (-0.048) & \cellcolor{orange!20}0.074 & \cellcolor{orange!20}0.068 & \cellcolor{orange!20}0.049 & 0.295 & 0.210 \\
Herding                       & 0.177 & 0.117 & 0.130 & 0.093 & 0.332 & 0.245 \\
\quad + Ours (Avg)          & 0.131 (-0.046) & 0.078 & 0.076 & 0.056 & 0.282 & 0.199 \\
\quad + Ours (Unc)          & 0.135 (-0.042) & \cellcolor{orange!20}0.074 & 0.069 & \cellcolor{orange!20}0.049 & 0.302 & 0.210 \\
K-Center Greedy               & 0.172 & 0.111 & 0.123 & 0.087 & 0.331 & 0.239 \\
\quad + Ours (Avg)  & \cellcolor{orange!20}0.129 (-0.043) & 0.076 & 0.073 & 0.053 & \cellcolor{orange!20}0.281 & \cellcolor{orange!20}0.203 \\
\quad + Ours (Unc)  & 0.131 (-0.042) & \cellcolor{orange!20}0.074 & 0.069 & \cellcolor{orange!20}0.049 & 0.291 & 0.208 \\
\hline
\end{tabular}}
\label{tab:coreset_with_ours_diff_ratios}
\end{table*}

\end{document}